\documentclass[10pt,twocolumn,letterpaper]{article}

\usepackage{iccv}
\usepackage{times}
\usepackage{epsfig}
\usepackage{graphicx}
\usepackage{amsmath}
\usepackage{amssymb}
\usepackage{multirow}
\usepackage{pifont}
\usepackage{bm}
\usepackage{subcaption}
\usepackage{booktabs}
\usepackage{colortbl}
\usepackage[dvipsnames]{xcolor}
\usepackage [english]{babel}
\usepackage [autostyle, english = american]{csquotes}
\usepackage{comment}
\MakeOuterQuote{"}

\newcommand{\redcmd}[1]{\textcolor{BrickRed}{#1}}
\newcommand{\greencmd}[1]{\textcolor{ForestGreen}{#1}}

\newcommand{\ooc}{out-of-context}
\newcommand{\Ooc}{Out-of-Context}
\newcommand{\cic}{conflicting-image-captions}
\newcommand{\Cic}{Conflicting-Image-Captions}

\newcommand{\COUNTIMAGES}{200K}
\newcommand{\COUNTCAPTIONS}{450K}
\newcommand{\COUNTVALIDATION}{1700}
\newcommand{\cmark}{\ding{51}}%
\newcommand{\xmark}{\ding{55}}%

\definecolor{ashgrey}{rgb}{0.7, 0.75, 0.71}
\makeatletter
\newcommand\footnoteref[1]{\protected@xdef\@thefnmark{\ref{#1}}\@footnotemark}
\makeatother

\usepackage[breaklinks=true,bookmarks=false]{hyperref}

\iccvfinalcopy 

\begin{document}

\title{\vspace{-1.9cm}COSMOS: Catching Out-of-Context \\ Misinformation with Self-Supervised Learning}


\author{%
Shivangi Aneja$^{1}$~~~~~~~~Chris Bregler$^{2}$~~~~~~~~Matthias Nie{\ss}ner$^{1}$ \vspace{0.2cm}\\
$^{1}$Technical University of Munich~~~~~~~~$^{2}$Google AI
}


\twocolumn[{%
\renewcommand\twocolumn[1][]{#1}%
\maketitle
\begin{center}
    \centering
    \includegraphics[width=1.0\linewidth]{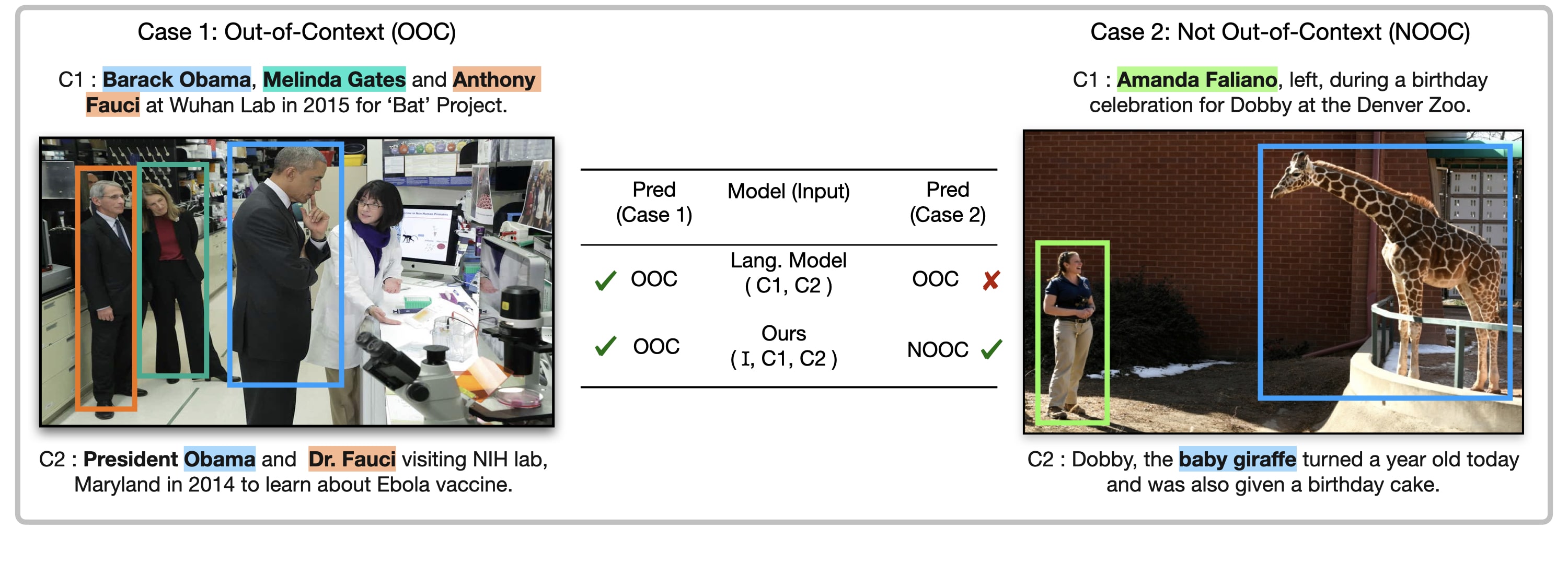}
    \vspace{-1.0cm}
    \captionof{figure}{
    Our method takes as input an image and two captions from different sources, and we predict whether the image has been used out of context or not. We show that it is critical to the task to ground the captions w.r.t. image, and it is insufficient to consider only the captions; e.g., a language-only model would incorrectly classify the right image to be out of context. To this end, we propose a new  self-supervised learning strategy allowing to make fairly accurate out-of-context predictions.
    %
%
%
    }
    \label{fig:teaser}
\end{center}%
}]


\begin{abstract}
Despite the recent attention to DeepFakes, one of the most prevalent ways to mislead audiences on social media is the use of unaltered images in a new but false context.
To address these challenges and support fact-checkers, we propose a new method that automatically detects \ooc\ image and text pairs.
Our key insight is to leverage grounding of image with text to distinguish out-of-context scenarios that cannot be disambiguated with language alone.
We propose a self-supervised training strategy where we only need a set of captioned images.
At train time, our method learns to selectively align individual objects in an image with textual claims, without explicit supervision.
At test time, we check if both captions correspond to same object(s) in the image but are semantically different, which allows us to make fairly accurate \ooc\ predictions. Our method achieves 85\% \ooc{} detection accuracy.
To facilitate benchmarking of this task, we create a large-scale dataset of \COUNTIMAGES{} images  with \COUNTCAPTIONS{} textual captions from a variety of news websites, blogs, and social media posts.
The dataset and source code is publicly available here\footnote{\url{https://shivangi-aneja.github.io/projects/cosmos/}}.
%
%
\end{abstract}

\begin{figure*}[ht]
\begin{center}
\vspace{-0.5cm}
\includegraphics[width=\linewidth]{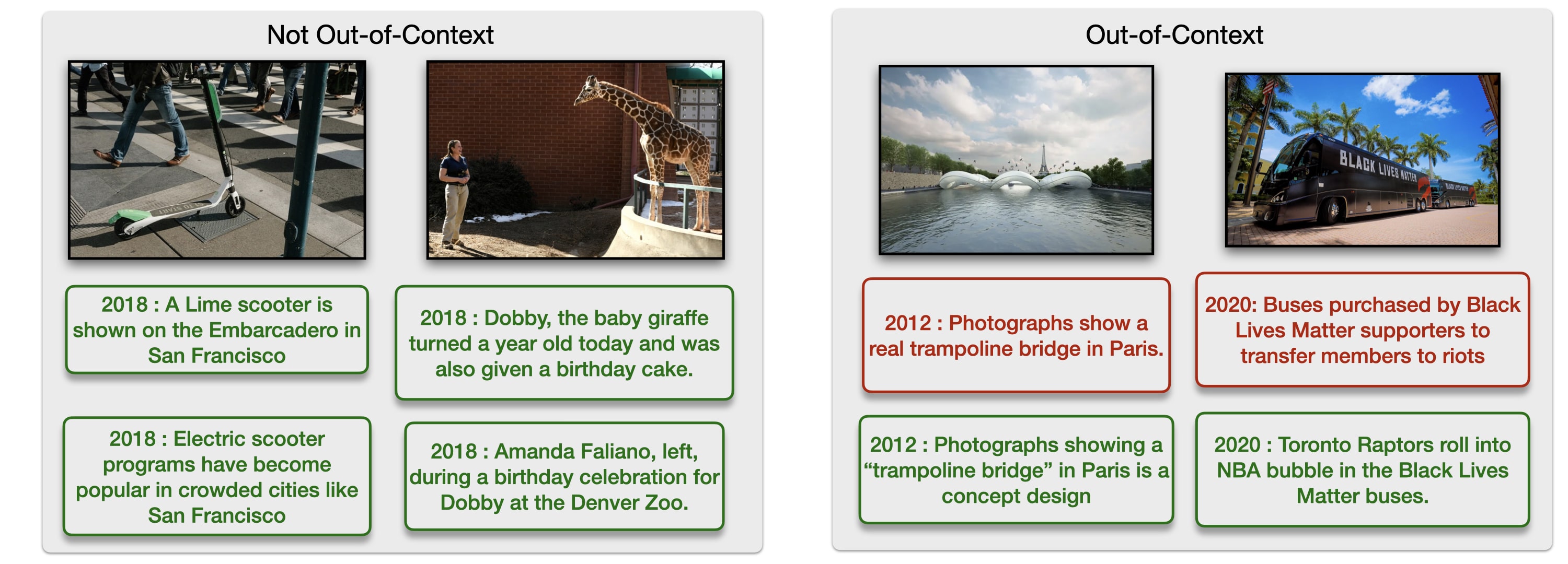}
\end{center}
\vspace{-0.5cm}
   \caption{Examples from our dataset where images from social media and online news were used out of context (right) and those which were not (left). (\redcmd{Red}) denotes false captions and (\greencmd{green}) shows the true captions along with year published.
   }
\label{fig:factcheck_samples}
\end{figure*}

\section{Introduction}

In recent years, the computer vision community as well as the general public have focused on new misuses of media manipulations such as DeepFakes~\cite{df_github,fs_gan,df_cf_19,petrov2020deepfacelab} and how they aid the spread of misinformation in news and social media platforms.
At the same time, researchers have developed impressive media forensic methods to automatically detect these manipulations~\cite{ff_dataset,Nguyen_2019,face_warping,deepfake_inconsistent_head_pose,Zhou_2017,cozzolino2018forensictransfer,mesonet,agarwal_protecting_2019,li2020face,verdoliva2020media,aneja2020generalized,cozzolino2020idreveal}. 
However, despite the importance of DeepFakes and other visual manipulation methods, one of the most prevalent ways to mislead audiences is the use of unaltered images in a new but false or misleading context~\cite{ooc_blog_20}.
Fact checkers refer to this as out-of-context use of images, where an image appears on with two (or even more) online sources with different and contradictory captions.

%

The danger of out-of-context images is that little technical expertise is required, as one can simply take an image from a different event and create highly convincing but potentially misleading messages. 
At the same time, it is extremely challenging to detect misinformation based on \ooc{} images given that the visual content by itself is not manipulated; only the image-text combination creates misleading or false information.
In order to detect these \ooc\ images, several online fact-checking initiatives have been launched by news rooms and independent organizations, most of them currently being part of the International Factchecking Network \cite{ifcn_20}.
However, they all heavily rely on manual human efforts to verify each post factually, and to determine if a fact-checking claim should be labelled as ``out-of-context'' or not. 
Thus, automated techniques can aid and speed up verification of potentially false claims for fact checkers.

Seminal works along these lines focus on predicting the veracity of a claim based on certain evidence like subject, context, social network spread, prior history, etc.~\cite{wang-2017-liar,thorne-etal-2018-fever}.
However, these methods are limited only to the linguistics domain, focusing on textual metadata to predict the factuality of the claim. 
In particular, language-only analysis cannot accurately identify many out-of-context scenarios, as shown in Figure~\ref{fig:teaser} -- the grounding of which objects in an image the language refers to is essential towards understanding whether there is an out-of-context situation.

\begin{figure*}[t!]
\subfloat{\includegraphics[width=0.70\textwidth]{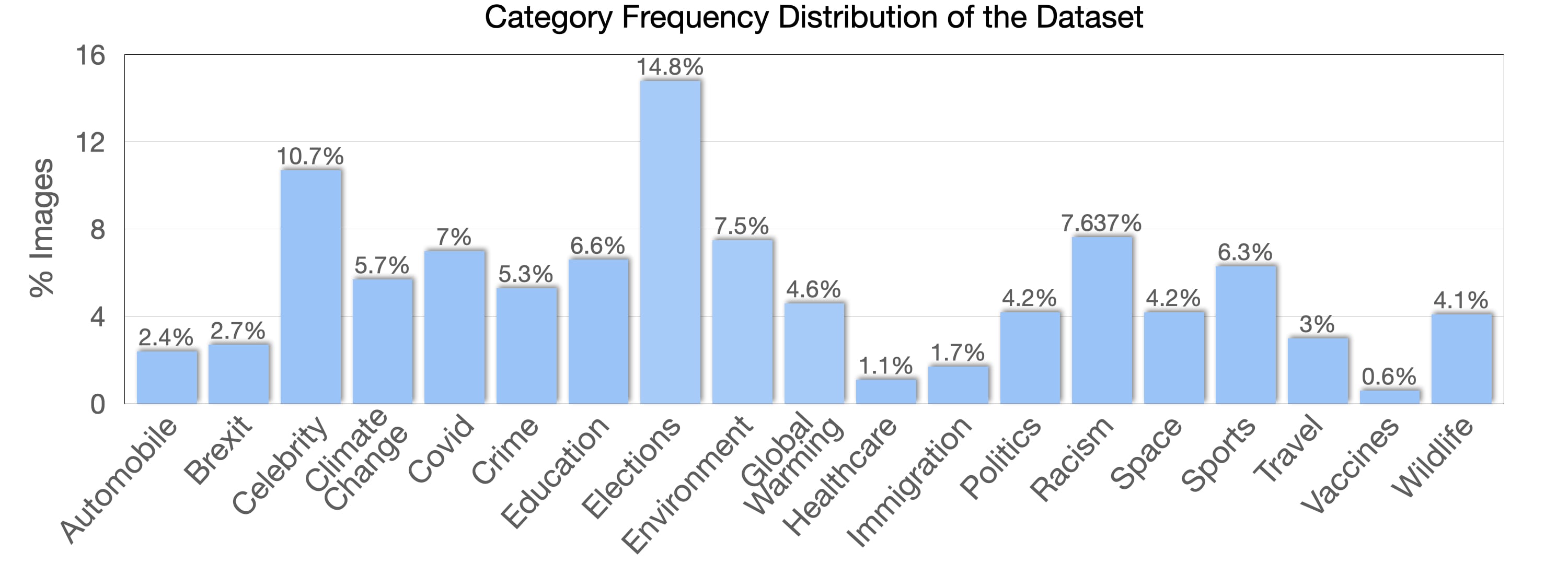}}
\hfill
   \subfloat{
   \includegraphics[width=0.30\textwidth]{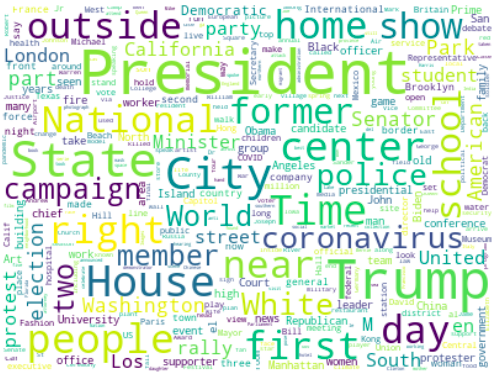}}
\caption{High-level overview of our dataset: (left) category-wise frequency distribution of the images; (right) word cloud representation of captions and claims from the dataset.}
\label{fig:dataset_topics}
\end{figure*}

We define \ooc{} use of images as presenting the image as an evidence of untrue or unrelated event(s). If the two captions refer to same object in the image, but are semantically different, i.e. correspond to different events, then it indicates \ooc{} use of image. 
However, if the captions correspond to the same event irrespective of the object(s) they captions describe, the it is defined as not-\ooc{}.

Note that a not-out-of-context scenario makes no conclusions regarding the veracity of the statements.
To automatically detect these cases, we propose a new data-driven method that takes an image and two text captions as input. As output, we predict whether the two captions referred to the image are \ooc\ or not.
 
The core idea of our method is a self-supervised training strategy where we only need captioned images; we do not require any explicit out-of-context annotations which would be potentially difficult to annotate in large numbers.
We can then establish the image captions from the data as matches, and random captions from other images as non-matches.
Using these matches vs non-matches as loss function, we are able to learn co-occurrence patterns of images with textual descriptions to determine whether the image appears to be \ooc\ with respect to textual claims.
During training, our method only learns to selectively align individual objects in an image with textual claims, without explicit \ooc{} supervision.
At test time, we are able to correlate these alignment predictions between the two captions for the input image. If both texts correspond to same object but their meaning is semantically different, we infer that the image is used \ooc{}.

In order to train our approach, we create a large-scale dataset of over \COUNTIMAGES{} images with their corresponding \COUNTCAPTIONS{} textual captions (some images appear with various captions, although not a necessary requirement) from a variety of news websites, blogs, and social media posts.
We further manually annotated a subset of \COUNTVALIDATION{} triplet pairs (an image and 2 captions) for benchmarking purposes only.
In the end, our method significantly improves over alternatives, reaching over 85\% detection accuracy.

\vspace{0.1cm}
In summary, our contributions are as follows:
\vspace{-0.1cm}
\begin{itemize}
  \item This paper proposes the first automated method to detect out-of-context use of images.
  \item We introduce a self-supervised training strategy for accurate out-of-context prediction while only using captioned images.
  \item We created a large dataset of \COUNTIMAGES{} images with \COUNTCAPTIONS{} corresponding text captions from a variety of news websites, blogs, and social media posts.
\end{itemize}

\section{Related Work}

\paragraph{Fake News \& Rumor Detection.}
Fake news and rumor detection methods have a long history~\cite{Qazvinian2011RumorHI,Kwon2013ProminentFO,Liu2015RealtimeRD,Ma2016DetectingRF,Zhao2017SpottingIB,Ruchansky2017CSIAH,Ma2018DetectRA,Ma2018RumorDO} and with the advent of deep learning, these techniques have accelerated in progress.
Most fake news and rumor detection methods focus on posts shared on microblogging platforms like Twitter. 
Kwon et al.~\cite{Kwon2013ProminentFO} analyzed structural, temporal, and linguistic aspects of the user tweets and modelled them using SVM to detect the spread of rumors. 
Liu et al.~\cite{Liu2015RealtimeRD} proposed an algorithm to debunk rumors in real-time. 
Ma et al.~\cite{Ma2018RumorDO} examined propagation patterns in tweets and applied tree-structured recursive neural networks for rumor representation learning and classification.
Tan et al.~\cite{tan2020detecting} detect neural fake news by exploiting visual and semantic inconsistencies in the news article.
 
\paragraph{Automated Fact-Checking.}
In recent years, several automated fact-checking techniques~\cite{wang-2017-liar,thorne-etal-2018-fever,Hasanain2019OverviewOT,atanasova-etal-2020-generating,ostrowski2020multihop,Atanasova2019AutomaticFU,vasileva-etal-2019-takes} have been developed to reduce the manual fact-checking overhead. 
For instance, Wang et al.~\cite{wang-2017-liar} created a dataset of short statements from several political speeches and designed a technique to detect fake claims by analyzing linguistic patterns in the speeches. 
Vasileva et al.~\cite{vasileva-etal-2019-takes} proposed a technique to estimate check-worthiness of claims from political debates. 
Atanasova et al.~\cite{atanasova-etal-2020-generating} propose a multi-task learning technique to classify veracity of claim and generate fact-checked explanations at the same time based on ruling comments. 

\paragraph{Verifying Claims about Images.}
Both fake news detection and automated fact-checking techniques are extremely important to combat the spread of misinformation and there are ample methods available to tackle this challenge. 
However, these methods target only textual claims and therefore cannot be directly applied to claims about images. To detect the increasing number of false claims about images, few methods~\cite{zhiwei_17,fauxbuster_2018, Shang2020FauxWardAG, Wang_eann_18,zlatkova-etal-2019-fact,Khattar2019MVAEMV} have been proposed recently. 
For instance, Khattar et al.~\cite{Khattar2019MVAEMV} learn a variational autoencoder based on shared embedding space (textual and visual) with binary classifier to detect fake news. 
Jin et al.~\cite{zhiwei_17} use attention-based RNNs to fuse multiple modalities to detect rumors/fake claims. 
Since these techniques are supervised in nature, they require large amounts of labelled data, which is difficult to obtain, especially for false claims. We, however, propose a self-supervised method to achieve the goal.

\section{Out-of-Context Detection Dataset}
Our dataset is based on images from news articles and social media posts.  
We gather images from a wide variety of articles (Fig.~\ref{fig:dataset_topics}), with special focus on topics where misinformation spread is prominent. 

\subsection{Dataset Collection}
We gathered our dataset from two primary sources, \textit{news websites}\footnote{\label{note1}New York Times, CNN, Reuters, ABC, PBS, NBCLA, AP News, Sky News, Telegraph, Time, DenverPost, Washington Post, CBC News, Guardian, Herald Sun, Independent, CS Gazette, BBC} and \textit{fact-checking websites}. 
We collect our dataset in two steps: 
(1) First, using publicly available news channel APIs, such as from the New York Times~\cite{nyt_api_20}, we scraped images along with the corresponding captions. 
(2) We then reverse-searched these images using Google's Cloud Vision API to find other contexts in which the image is shared. 
Note that the second step is not required for training, but we collect these captions for benchmarking as well as increased dataset size.
%
%
Thus, we obtain captioned images that we can use to train our models.
Note that we do not consider digitally-altered/fake images; our focus here is to detect misuse of real photographs. 
Finally, we currently aim to detect \cic\ in the English language only.

\subsection{Data Sources \& Statistics}
We obtained our images primarily from \textit{news channels} and a fact-checking website (\textit{Snopes}). 
We scraped images on a wide variety of topics ranging from \textit{politics, climate change, environment, etc} (see Fig.~\ref{fig:dataset_topics}). 
For images scraped from \textit{New York Times}, we used publicly available Article Search developer API~\cite{nyt_api_20}, and for other new sources, we wrote our custom scrapers.
For images from news channels, we scraped corresponding image captions from $<$\textit{figcaption}$>$ tag and \textit{alt text} attribute, and for Snopes, we scraped text written in the $<$\textit{Claim}$>$ header, under the \textit{Fact Checks} section of the website. 
In total, we obtain \COUNTIMAGES{} train images and \COUNTVALIDATION{} test images; see Tab.~\ref{tab:Dataset Split and stats}.

\begin{table}[bht!]
\begin{center}
\begin{tabular}{c|c|c|c}
\toprule
Split & Primary & No. of  & Context  \\
 & Source & Images & Annotation\\
\toprule
Train & News Outlets\footnoteref{note1} & 160K & \xmark\\
Val & News Outlets & 40K & \xmark\\
Test & News Outlets, Snopes & \COUNTVALIDATION & \cmark\\  
\hline
\end{tabular}
\end{center}
\vspace{-0.5cm}
\caption{Statistics of our out-of-context dataset.}
\label{tab:Dataset Split and stats}
\end{table}

\textbf{Train Set:} 
For training, we used images scraped from news websites. We consider several news sources\footnoteref{note1} to gather the images. 
In total, we gathered around \COUNTIMAGES{} images with \COUNTCAPTIONS{} captions, 20\% of which we use as the validation set.

\textbf{Test Set:} 
At test time, we use the images from the fact-checking website \textit{Snopes} along with news websites.
We collected \COUNTVALIDATION{} images with two captions per image.
We build an in-house annotation tool to manually annotate these pairs with \ooc\ labels. 
On average, it takes around 45 seconds to annotate every pair, and we spent 100 hours in total to collect and annotate the entire test set.
We ensured an equal distribution of both \ooc\ and not-\ooc\ images in the test split.

\section{Method}

We consider a dataset of captioned images, where images may have more than one associated caption; however, we do not have any mapping for the objects referenced by the captions nor labels for which captions are \ooc. 
We notice that in typical \ooc\ use of images, different captions often describe the same object(s) but with a different meaning.
For example, Fig.~\ref{fig:factcheck_samples} shows several fact-checked examples where the two captions mean something very different, but describe the same parts of the image. 
Our goal is to take advantage of these patterns to detect scenarios and identify images used \ooc. 

\begin{figure}[ht]
\begin{center}
\includegraphics[width=0.5\textwidth]{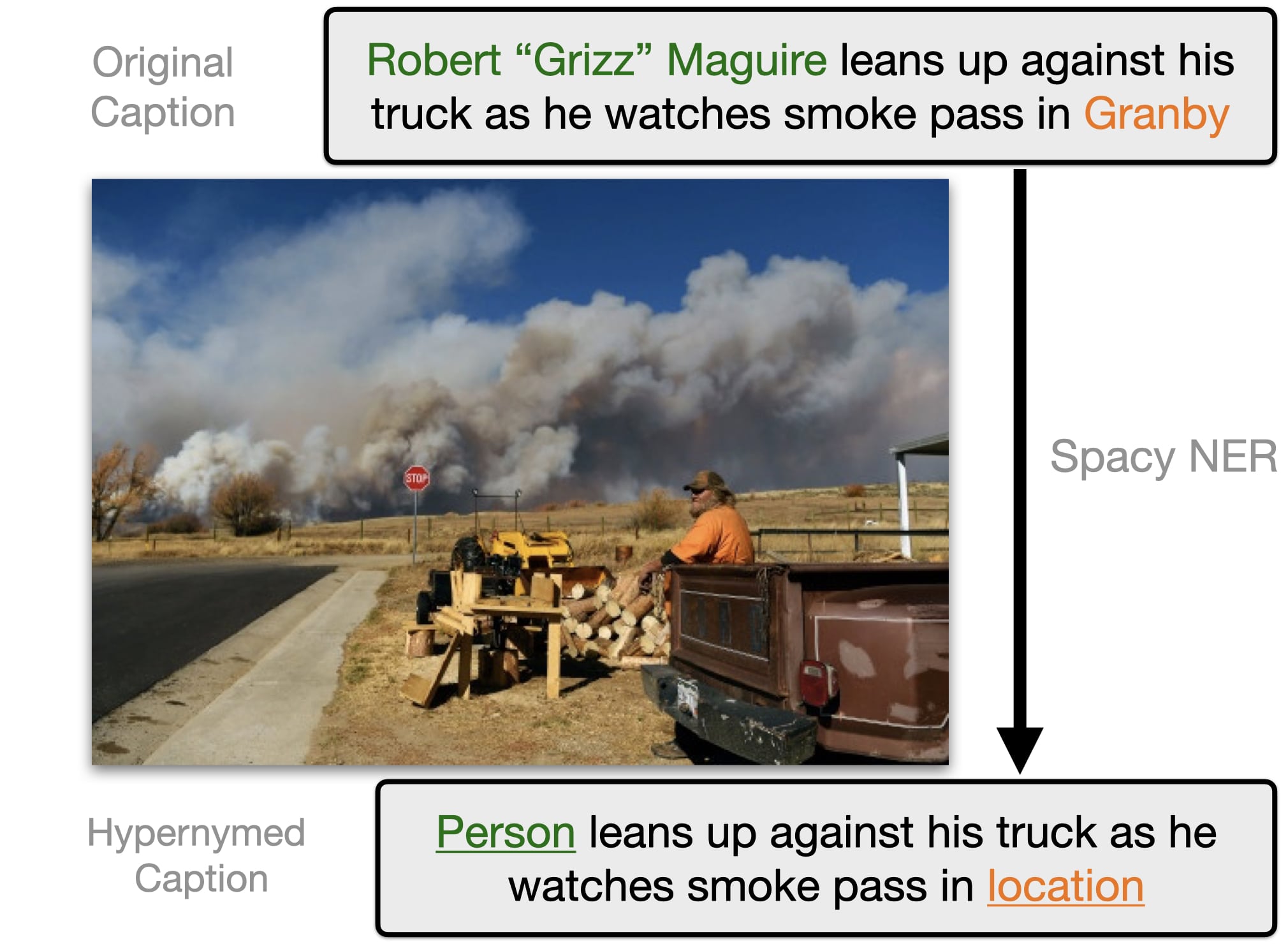}
\vspace{-0.6cm}
\end{center}
   \caption{Text Pre-processing: we pre-process captions to replace named entities in the image with their corresponding hypernyms. For instance, the person's name ``Robert Grizz Maguire" is replaced with the hypernym \textit{Person} and the town ``Granby" is replaced with the hypernym \textit{location}}
\label{fig:text_preprocess}
\end{figure}

\begin{figure*}[ht!]
\includegraphics[width=\linewidth]{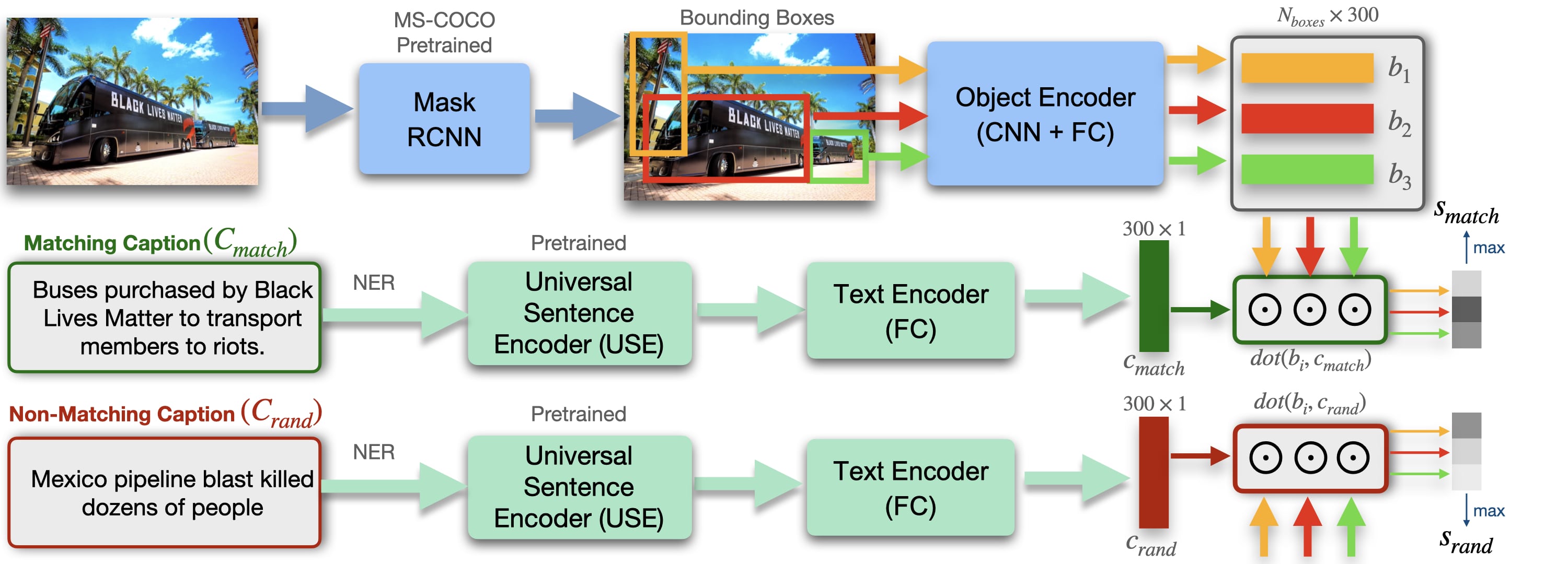}
\caption{Self-supervised training of our method. First, a Mask-RCNN~\cite{he2018mask} backbone detects up to 10 object boxes in the image whose regions are embedded through our Object Encoder, providing a fixed-size embedding for each object. In parallel, two captions -- one that appeared originally with the image $C_{match}$ (matching caption) and another caption sampled randomly $C_{rand}$ (non-matching caption) -- and encoded using the Universal Sentence Encoder model (USE)~\cite{cer-etal-2018-universal}. The sentences embeddings are then passed to a shared Text Encoder that embeds them in the same multi-modal space. Similarities between object-caption pairs are computed with inner products (grayscale indicating score magnitude) and finally reduced to scores following Eq.~\ref{eq:eq1}.}
\label{fig:train_method}
\end{figure*}

\subsection{Text Pre-processing}
Since the captions used in our dataset are scraped from news websites, most captions contain proper nouns such as a person's name, city/country, venues, etc., which is hard for a model to interpret and thus makes it difficult to learn correct grounding. Hence, we used Spacy Named Entity Recognizer (NER)\footnote{\label{note2}https://spacy.io/api/entityrecognizer} to replace named entities in all the captions with their hypernyms. An example is shown in Figure~\ref{fig:text_preprocess}. Note that we always input these cleaned and hypernymed captions to our model for all our experiments.

\subsection{Image-Text Matching Model (Training)}\label{sec:method_train}

The core of our method is a self-supervised training strategy leveraging co-occurrences of an image and its objects with several associated captions; i.e., we propose training an image and text based model based only on a set of captioned images.
We thus formulate a scoring function to align objects in the image with the caption. 
Intuitively, an image-caption pair should have a high matching score if visual correspondences for the caption are present in the image, and a low score if the caption is unrelated to the image.
To infer this correlation, we first use a pre-trained Mask-RCNN~\cite{he2018mask} to detect bounding boxes of objects in the image. 

For each detected bounding box, we then feed the corresponding object regions to our Object Encoder, which uses a ResNet-50~\cite{resnet_18} backbone from a pre-trained Mask-RCNN followed by RoIAlign,  average pooling, and two fully-connected layers.
As a a result, for each object, we obtain a 300-dimensional embedding vector.

In parallel, we consider the corresponding (pre-processed) image caption $C_{match}$, and sample a random caption from a different image in the dataset, $C_{rand}$. 
The captions are fed into a pre-trained sentence embedding model.
Specifically, we use the Universal Sentence Encoder (USE)~\cite{cer-etal-2018-universal}, which is based on a state-of-the-art transformer~\cite{vaswani2017attention} architecture and outputs a 512-dimensional vector. 
We then process this vector with our Text Encoder (ReLU followed by one FC layer), which outputs a 300-dimensional embedding vector for each caption (to match the dimension of the object embeddings).

We then compare the visual and language embeddings with a dot product between the $i$-th box embedding $b_i$ and the caption embedding $c$ as a measure of similarity between image region $i$ and caption $C$. 
The final image-caption score $S_{IC}$ is obtained through a max function:

\begin{equation}
 S_{IC} = \max_{i=1}^{N} (b_i^Tc), \hspace{2mm}     N=\# bboxes.
    \label{eq:eq1}
\end{equation}

Our objective is to obtain higher scores for aligned image-text pairs (i.e., if an image appeared with the text irrespective of the context) than misaligned image-text pairs  (i.e., some randomly-chosen text which did not appear with the image). 
We train the model with max-margin loss (Eq.~\ref{eq:eq2}) on the image-caption scores obtained above (Eq.~\ref{eq:eq1}). 
Note that we keep the weights of the Mask-RCNN~\cite{he2018mask} backbone and the USE~\cite{cer-etal-2018-universal} model frozen, using these models only for feature extraction.
\vspace{-0.1cm}
\begin{equation}
    \mathcal{L}  = \frac{1}{N} \sum_{i}^{N} \max(0, (S_{IC}^r - S_{IC}^m) + \text{margin}),
    \label{eq:eq2}
\end{equation}
where $S_{IC}^r$ denotes the image-caption score of the random caption and $S_{IC}^m$ the image-caption score for the matching caption.
We refer to this model as \textit{Image-Text Matching Model}; the training setup is visualized in Fig.~\ref{fig:train_method}. 
Note that during training, we do not aim to detect \ooc\ images, but rather learn accurate image-caption alignments.

\begin{figure*}[ht!]
\includegraphics[width=\linewidth]{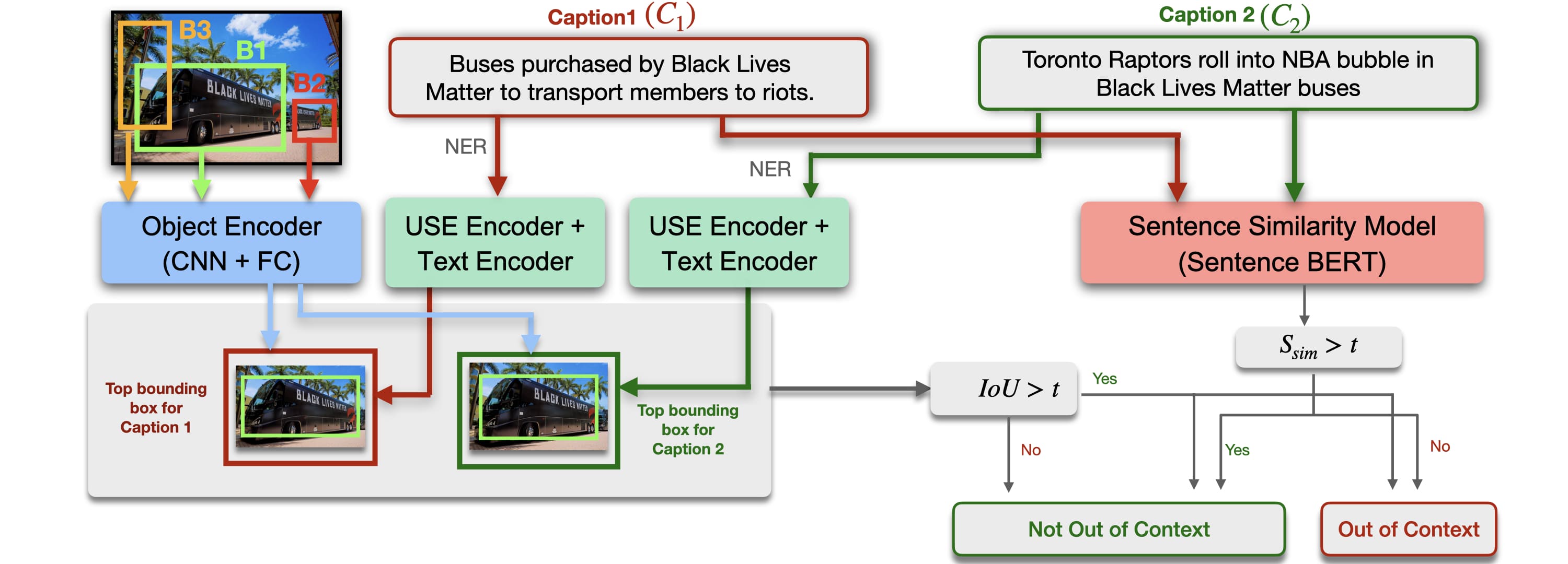}
\caption{Test time out-of-context detection. We take as input an image and two captions; we then use the trained Image-Text Matching model where we first pick the highest scoring object (based on Eq.~\ref{eq:eq1}) for both the captions. If the IoU between them $>$ threshold $t_i$, we infer that image regions overlap. If the image regions overlap, we compute textual overlap $S_{sim}$ with a pre-trained Sentence Similarity model SBert~\cite{sbert} and if $S_{sim} < t_s$, it implies that the two captions are semantically different, thus implying that the image is used out of context.}
\label{fig:test_time}
\end{figure*}

\subsection{\Ooc\ Detection Model (Test Time)}\label{sec:method_test}
The resulting Image-Text matching model obtained from training now provides an accurate representation of how likely a caption aligns with an image.
In addition, as we explicitly model the object-caption relationship, the max operator in Eq.~\ref{eq:eq1} implicitly gives a strong signal as to which object was selected to make that decision, thus providing spatial knowledge from the image.
At test time, we consider an image and two captions that it appeared with, which may or may not be semantically similar.
The Image-Caption1-Caption2 ($I,C_1,C_2$) triplet is used to predicts whether the image  was used \ooc\ with respect to the captions. 
Based on the evidence that \ooc\ pairs correspond to same object in the image (c.f. Fig.~\ref{fig:factcheck_samples}), we propose a simple rule to detect such images, i.e., if two captions align with same object(s) in the image, but semantically convey different meanings, then the image with its two captions is classified as \ooc{}.
More specifically, we make use of the pre-trained model as follows:

\smallskip
\noindent
(1) Using the Image-Text Matching model, we first compute the visual correspondences of the objects in the image for both captions. 
For each image-caption pair $\{I,C_j\}$, we choose the object box $B_{I,C_j}$ with the highest score $S_{I,C_j}$ by Eq.~\ref{eq:eq1} (strong alignment of caption with the object).

\smallskip
\noindent %
(2) We leverage a state-of-the-art SBERT~\cite{sbert} model that is trained on a Sentence Textual Similarity (STS) task. 
The SBERT model takes two captions $C_1, C_2$ as input and outputs a similarity score $S_{sim}$ in the range $[0,1]$ indicating semantic similarity between the two captions (higher score indicates same context):

\begin{equation}
    S_{sim} = \text{STS}(C1, C2)
    \label{eq:eq4}
\end{equation}

As a result, SBERT provides the semantic similarity between two captions, $S_{sim}$. 
In order to compute the visual mapping of the two captions with the image, we use the IoU overlap of the top bounding box for the two captions. 
We use thresholds $t_i=t_s=0.5$ for all our experiments, both for IoU overlap and text overlap.
If the visual overlap between image regions for the two captions is over a certain threshold $IoU(B_{I,C_1}, B_{I,C_2}) > t_i$
and the captions are semantically different ($S_{sim} < t_s$), we classify them as \ooc{} (OOC).
A detailed explanation is given in Fig.~\ref{fig:test_time} and is as follows:
\begin{equation}
    {\mathrm{OOC}}= 
\begin{cases}
    {\mathrm{True}}, \hspace{2mm} \text{if }\mathrm{IoU}\big(B_{IC_1}, B_{IC_2}\big) > t \hspace{2mm} \&  \\
    { \hspace{14mm} S_{sim}(C1, C2) < t}\\
    {\mathrm{False}}, \hspace{2mm} otherwise\\
\end{cases}
\label{eq:ooc_eq}
\end{equation}

\section{Results}

\subsection{Visual Grounding of Objects}\label{sec:refcoco}
\paragraph{Quantitative Results.}
Our model is trained in a self-supervised fashion only with matching and non-matching captions.
To quantitatively evaluate how well our model learns the visual grounding of objects, we use the RefCOCO dataset~\cite{yu2016modeling} which has ground truth associations of the captions with the object bounding boxes.
Note, however, that this evaluation is not our final task, but gives important insights into our model design.
We experiment with three different model settings:
(1) \textit{Full-Image}, where an image is fed as input to the model and directly combined with the text embedding. 
(2) \textit{Self-Attention}, where an image is fed as input to the model but combined with text using self-attention module. 
(3) \textit{Bbox}, where only the detected objects are fed as input to model instead of full image. 
For this experiment, we use the ILSVRC 2012-pre-trained ResNet-18~\cite{resnet_18} backbone to encode images and a one-layer LSTM model to encode text. 
%
Words are embedded using Glove~\cite{pennington2014glove} pre-trained embeddings.
Tab.~\ref{tab:tab1} shows that using object-level features (given by bounding boxes) gives the best performing model. 
This is unsurprising, as object regions can provide a richer feature representation for the entities in the caption compared to the full image; but we also significantly outperform a self-attention alternative.

\begin{table}[ht]
\begin{center}
\begin{tabular}{c|c|c}
\toprule
Img Features. & Object IoU & Match Acc. \\
\toprule
Bbox (GT) & 0.36 & 0.89 \\ \midrule 
Full-Image & 0.11 & 0.63 \\  
Self-Attention & 0.16 & 0.78 \\
Bbox (Pred) & \textbf{0.27} & \textbf{0.88}\\ 

\toprule
\end{tabular}
\end{center}
\vspace{-0.4cm}
\caption{
Ablation of different settings for visual grounding in our self-supervised training setting (no loss on IoU).
}
\label{tab:tab1}
\end{table}

\begin{figure*}[ht]
\begin{center}
\vspace{-0.2cm}
\subfloat{{\includegraphics[width=0.32\linewidth]{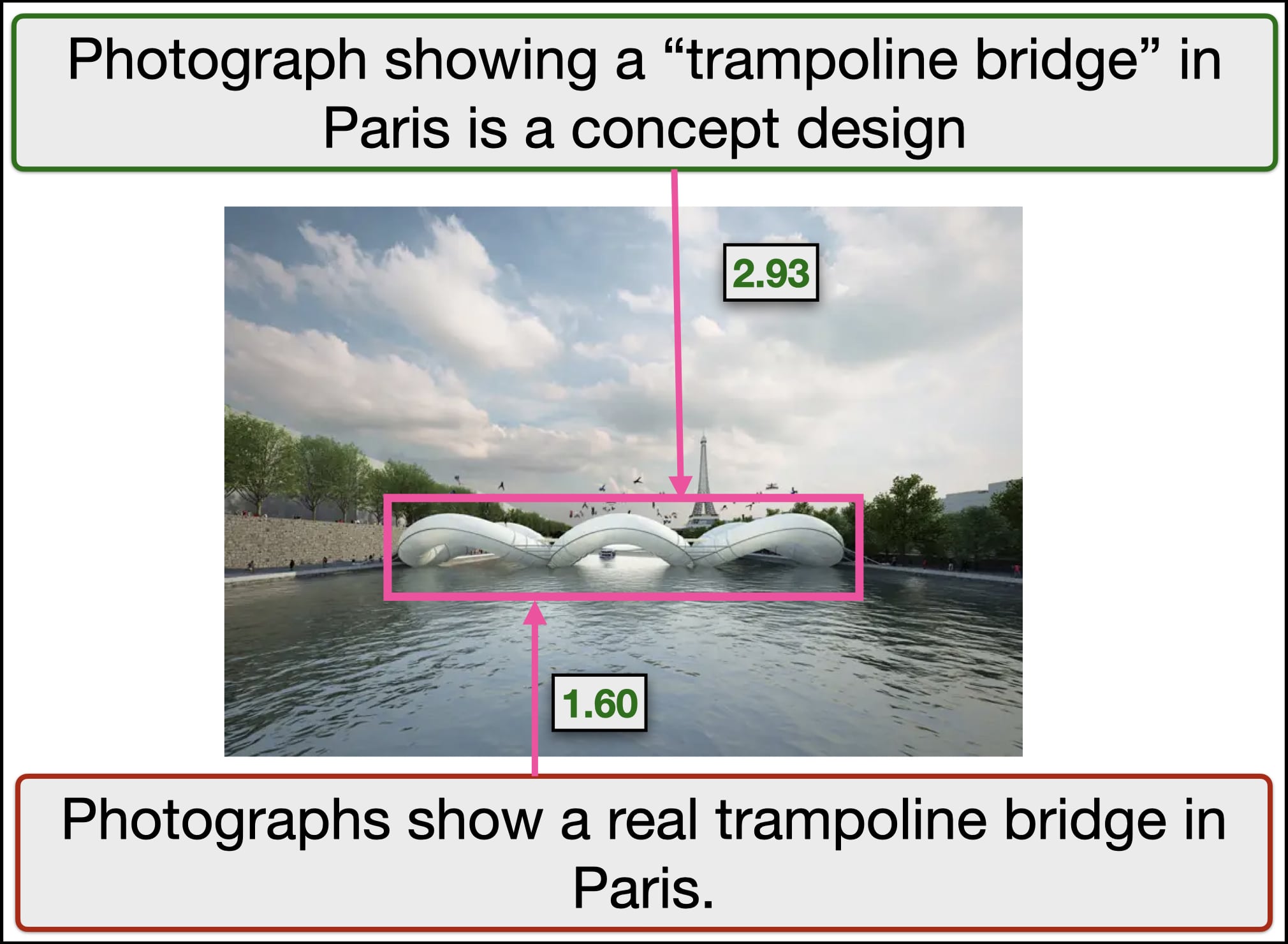} }}%
    \hfill
    \subfloat{{\includegraphics[width=0.32\linewidth]{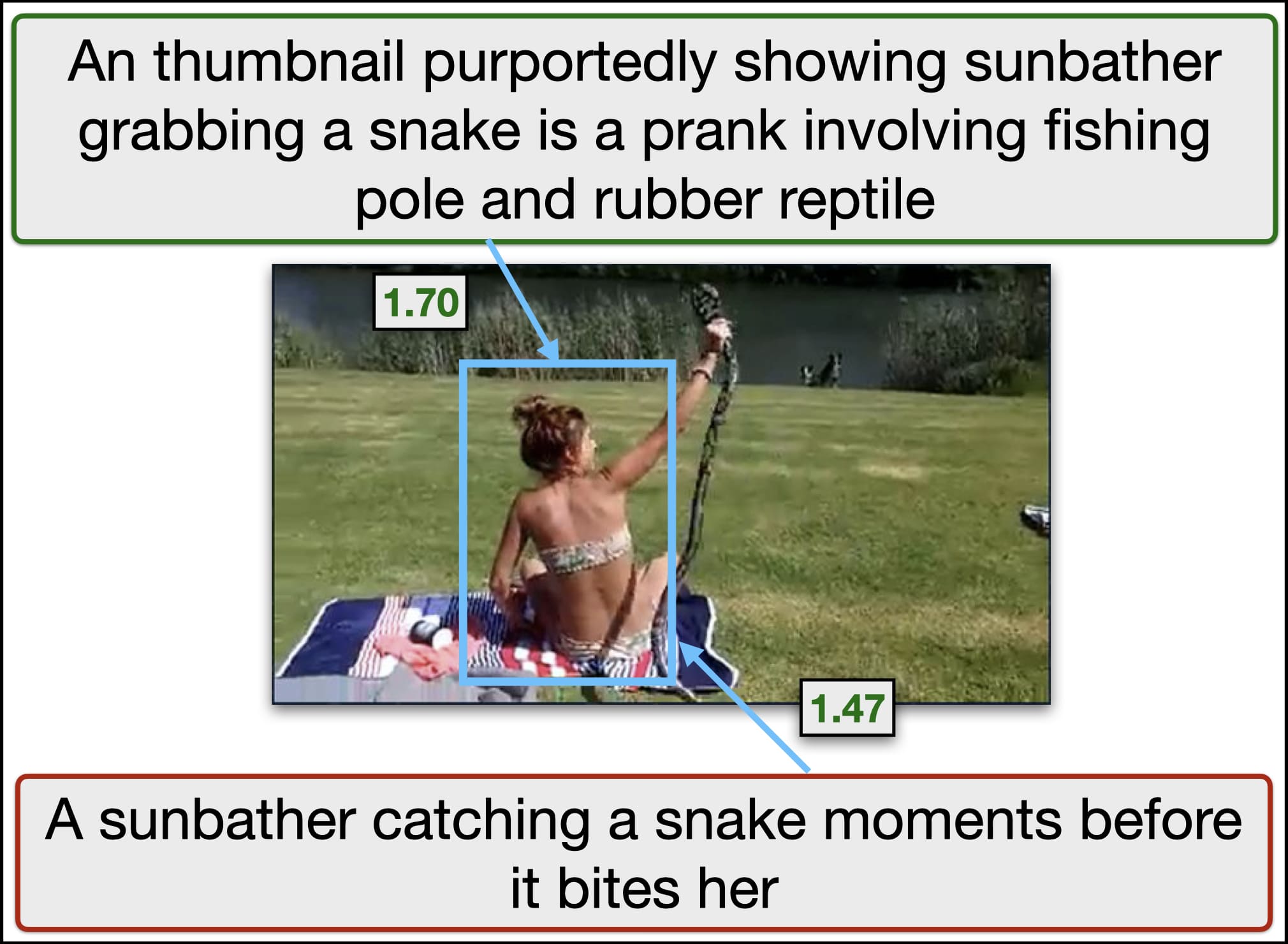} }}%
    \hfill
    \subfloat{{\includegraphics[width=0.32\linewidth]{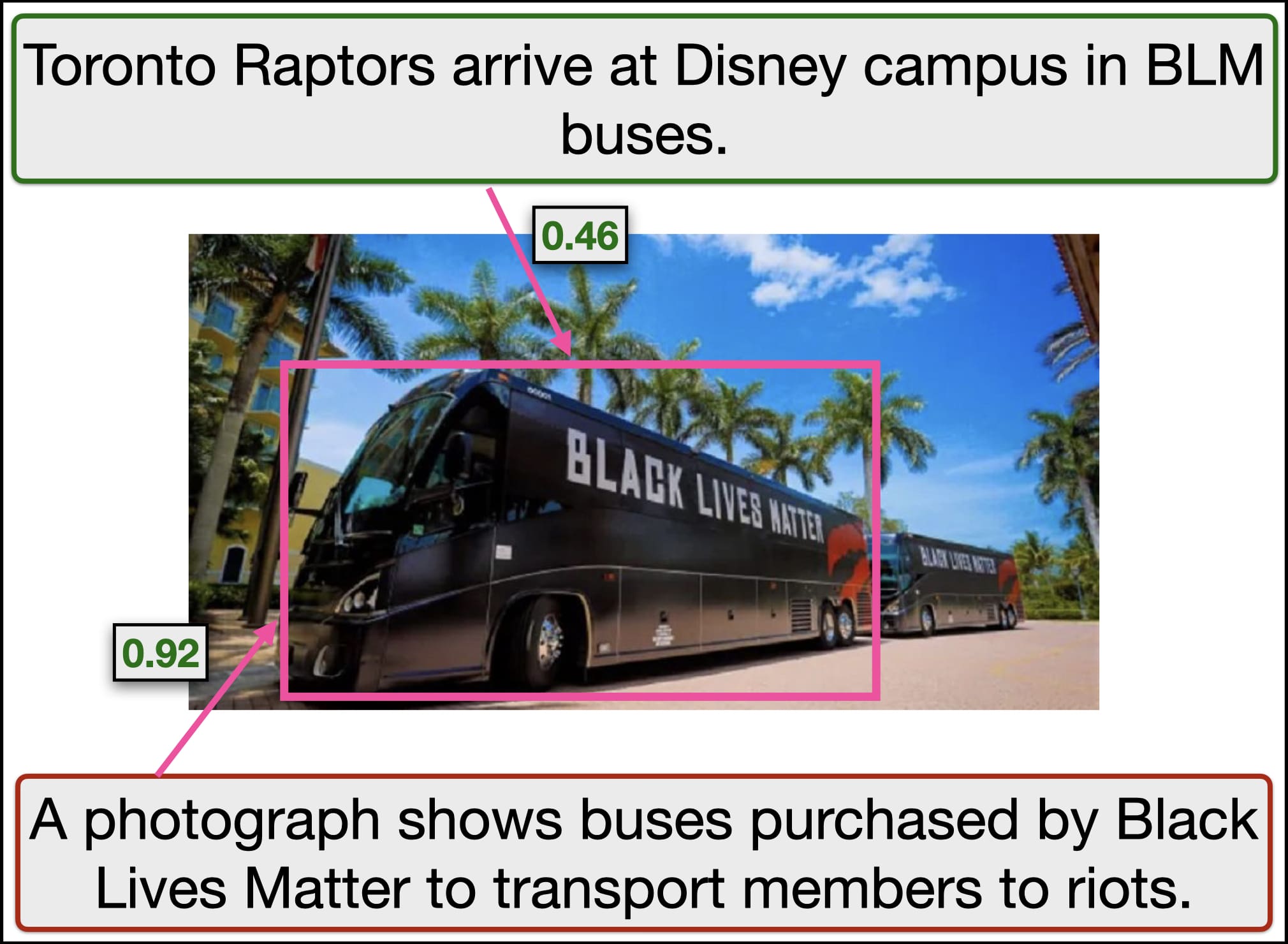} }}
    \end{center}
    
    \begin{center}
    \subfloat{{\includegraphics[width=0.32\linewidth]{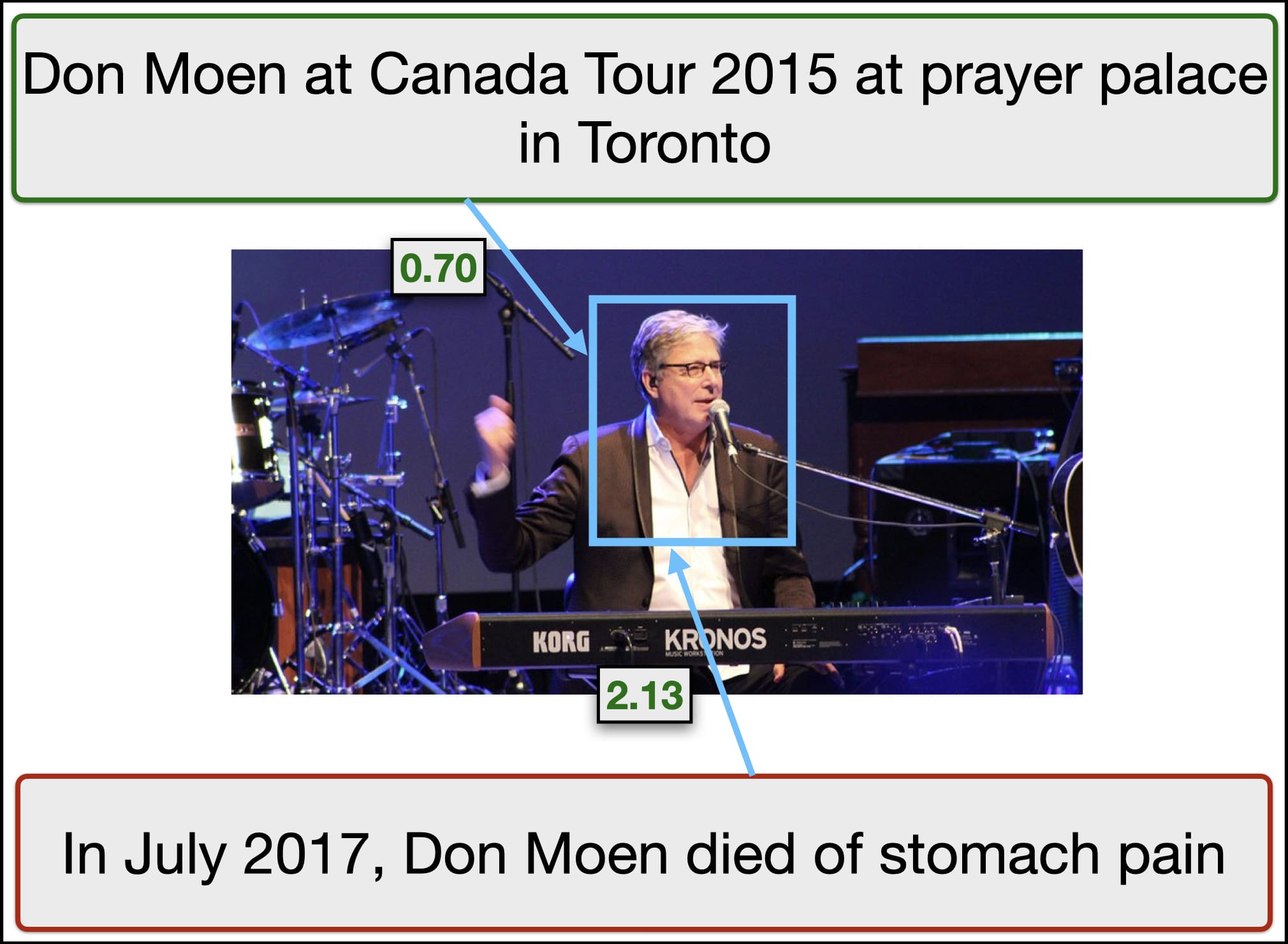} }}%
    \hfill
    \subfloat{{\includegraphics[width=0.32\linewidth]{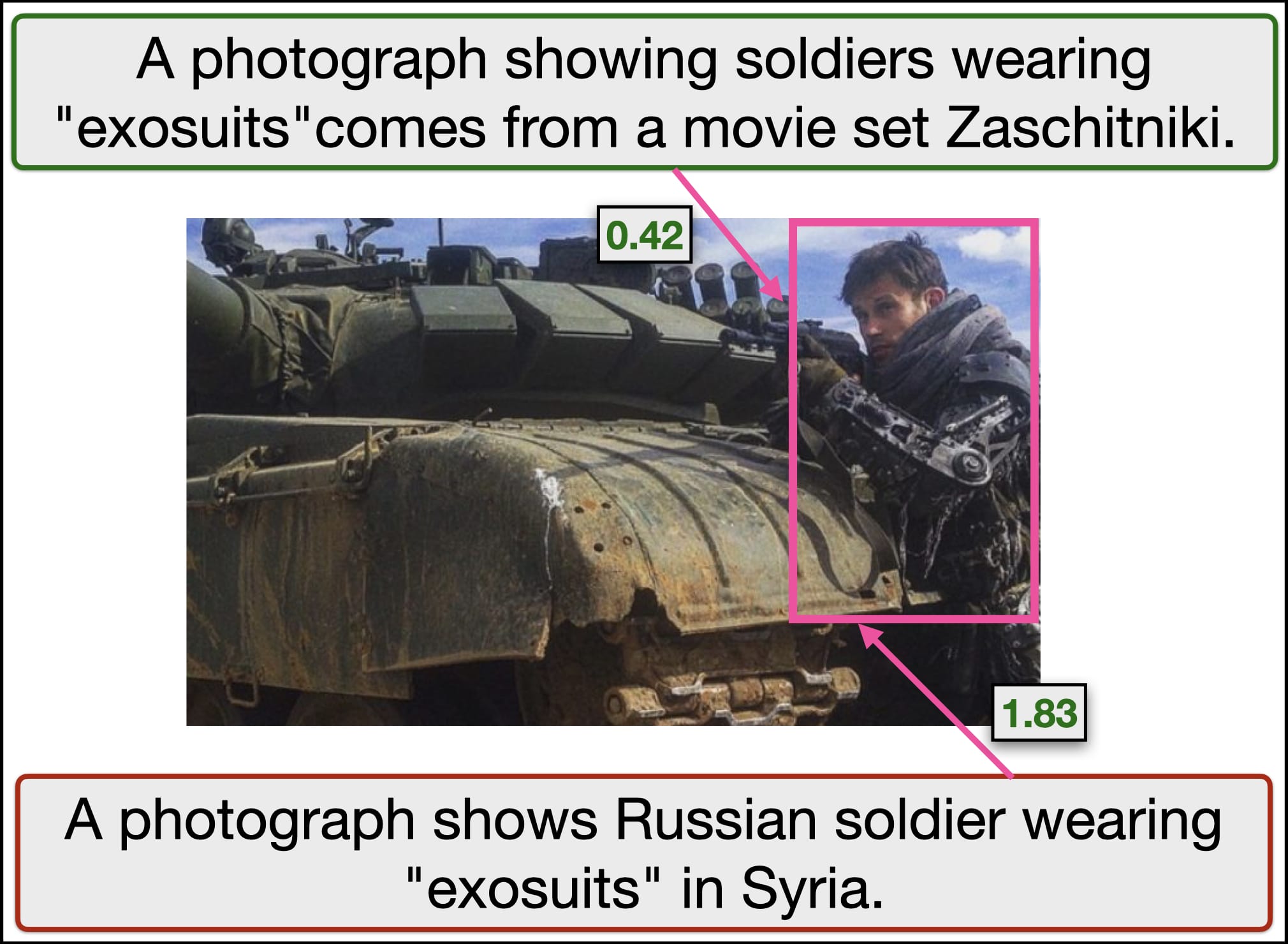} }}%
    \hfill
    \subfloat{{\includegraphics[width=0.32\linewidth]{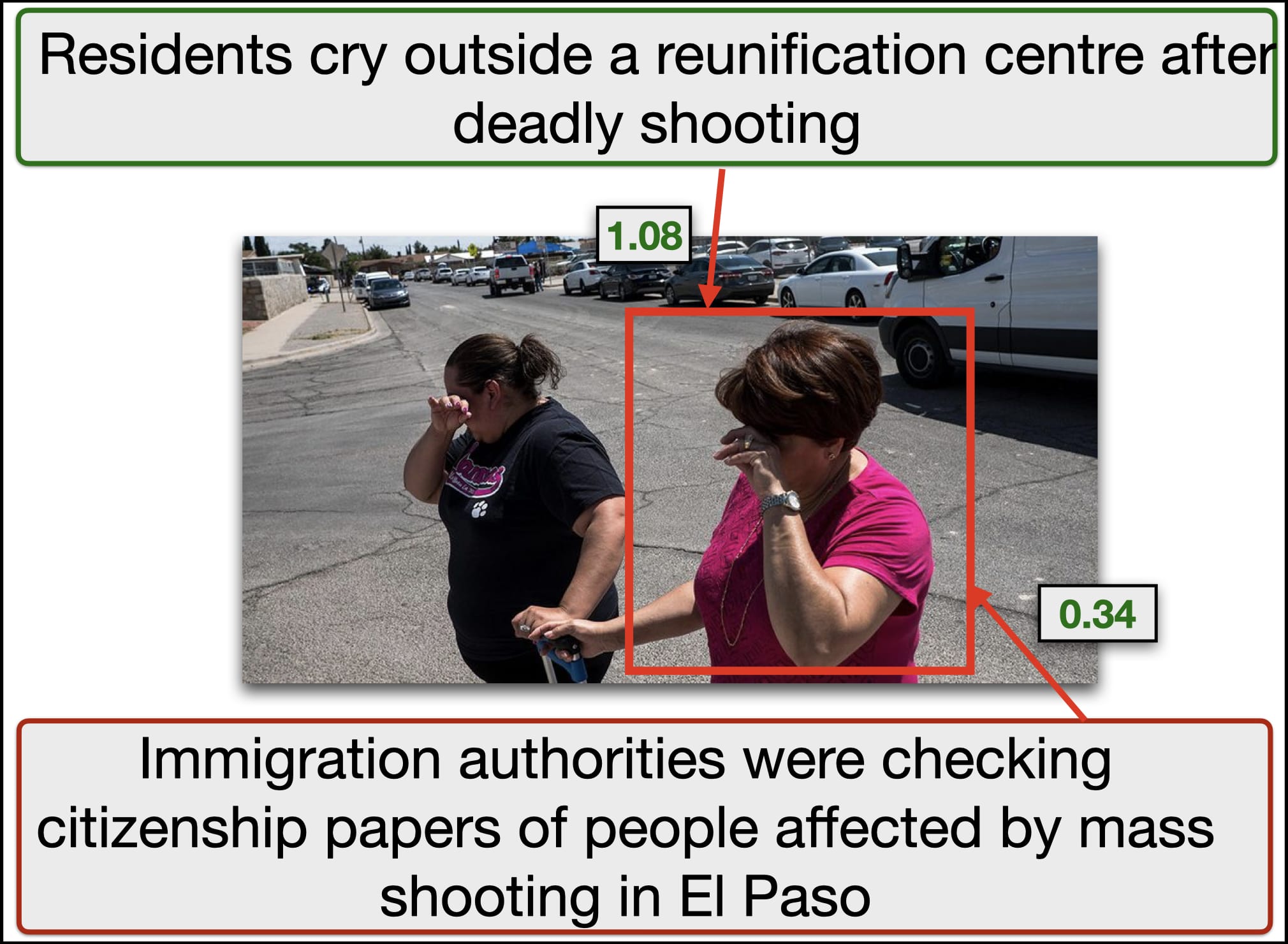} }}%
    \hfill
    \end{center}
    \begin{center}
    \subfloat{{\includegraphics[width=0.32\linewidth]{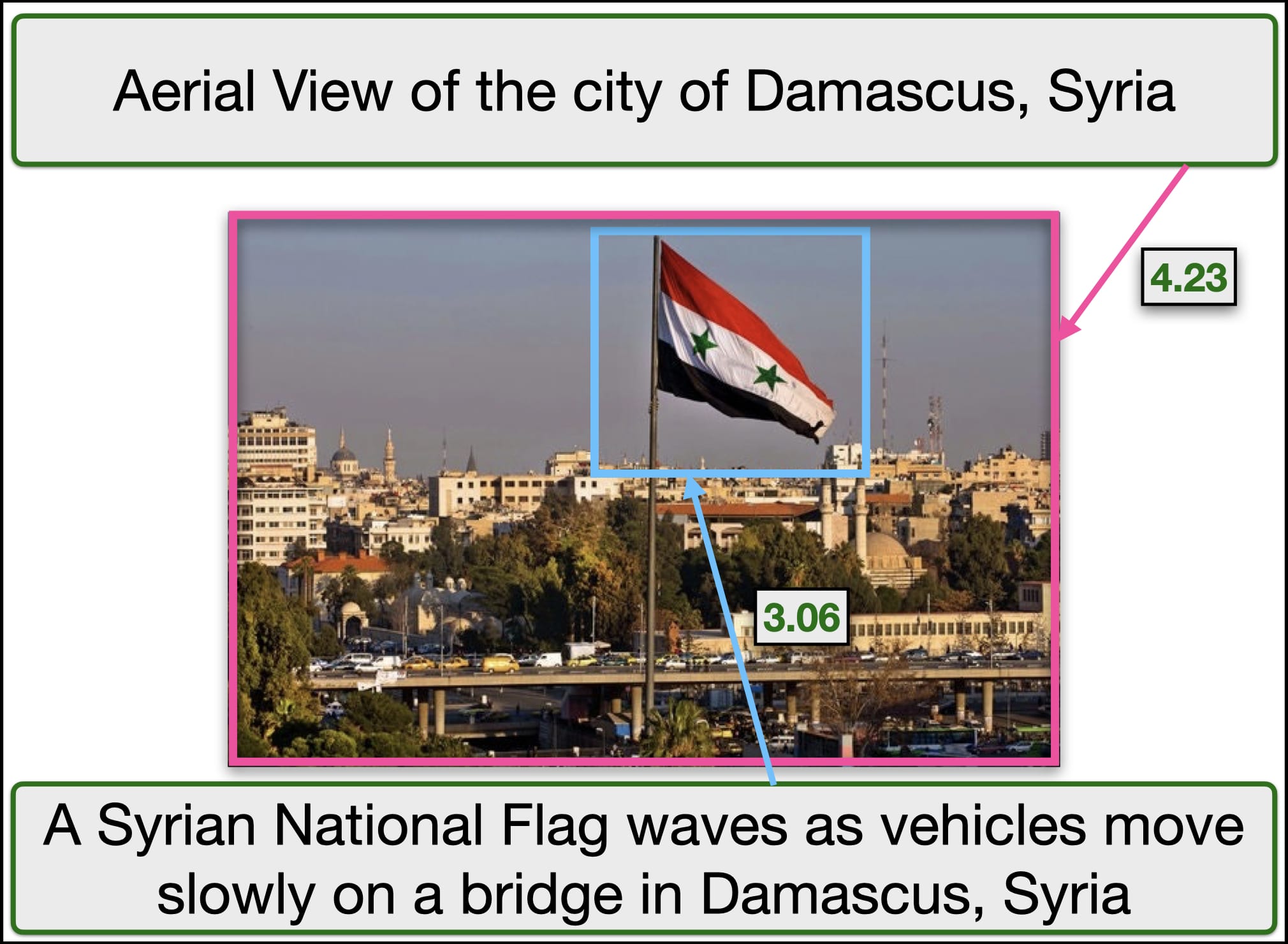} }}%
    \hfill
    \subfloat{{\includegraphics[width=0.32\linewidth]{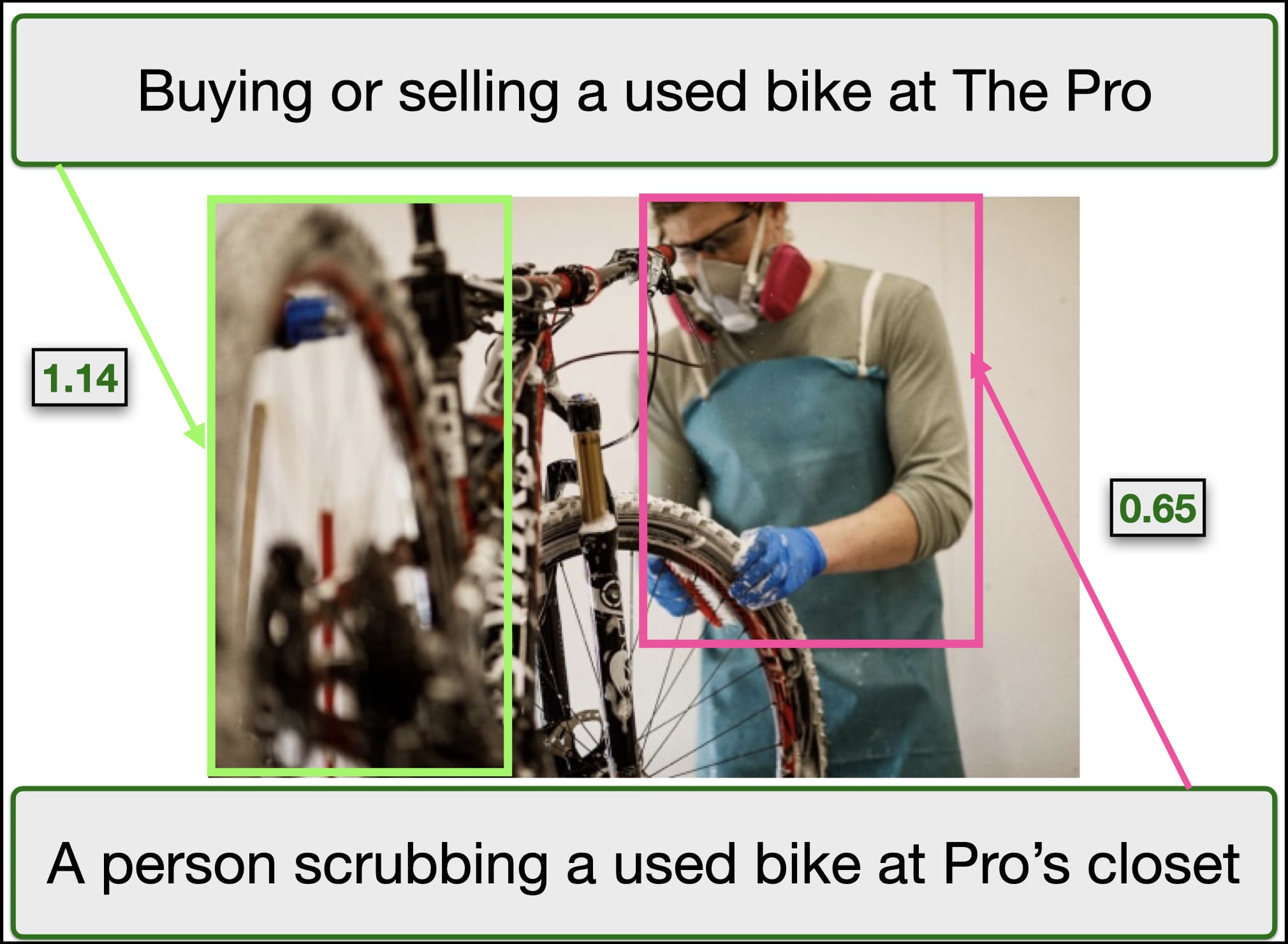} }}%
    \hfill
    \subfloat{{\includegraphics[width=0.32\linewidth]{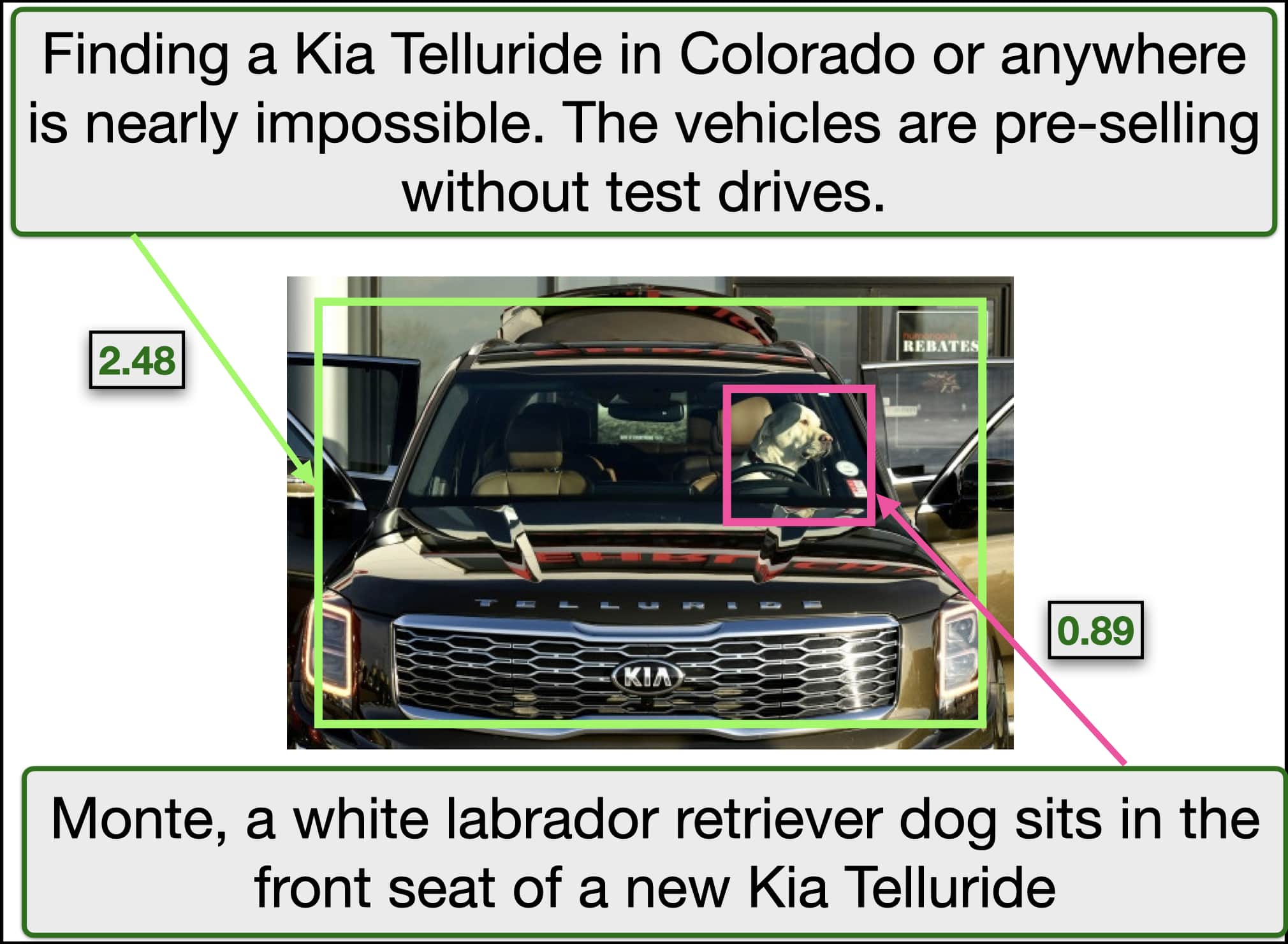} }}%
    \hfill
\end{center}
\begin{center}
    \subfloat{{\includegraphics[width=0.32\linewidth]{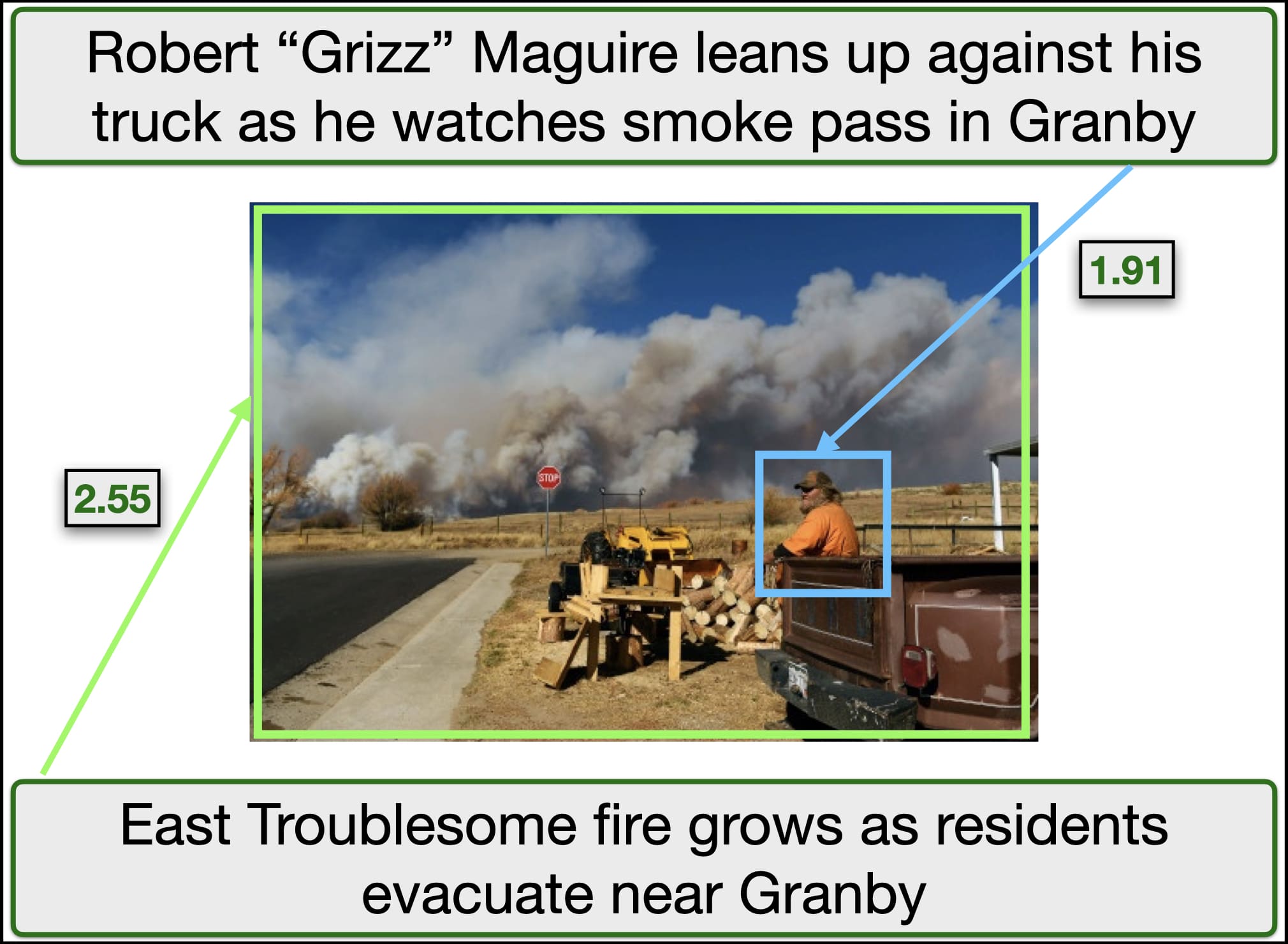} }}%
    \hfill
    \subfloat{{\includegraphics[width=0.32\linewidth]{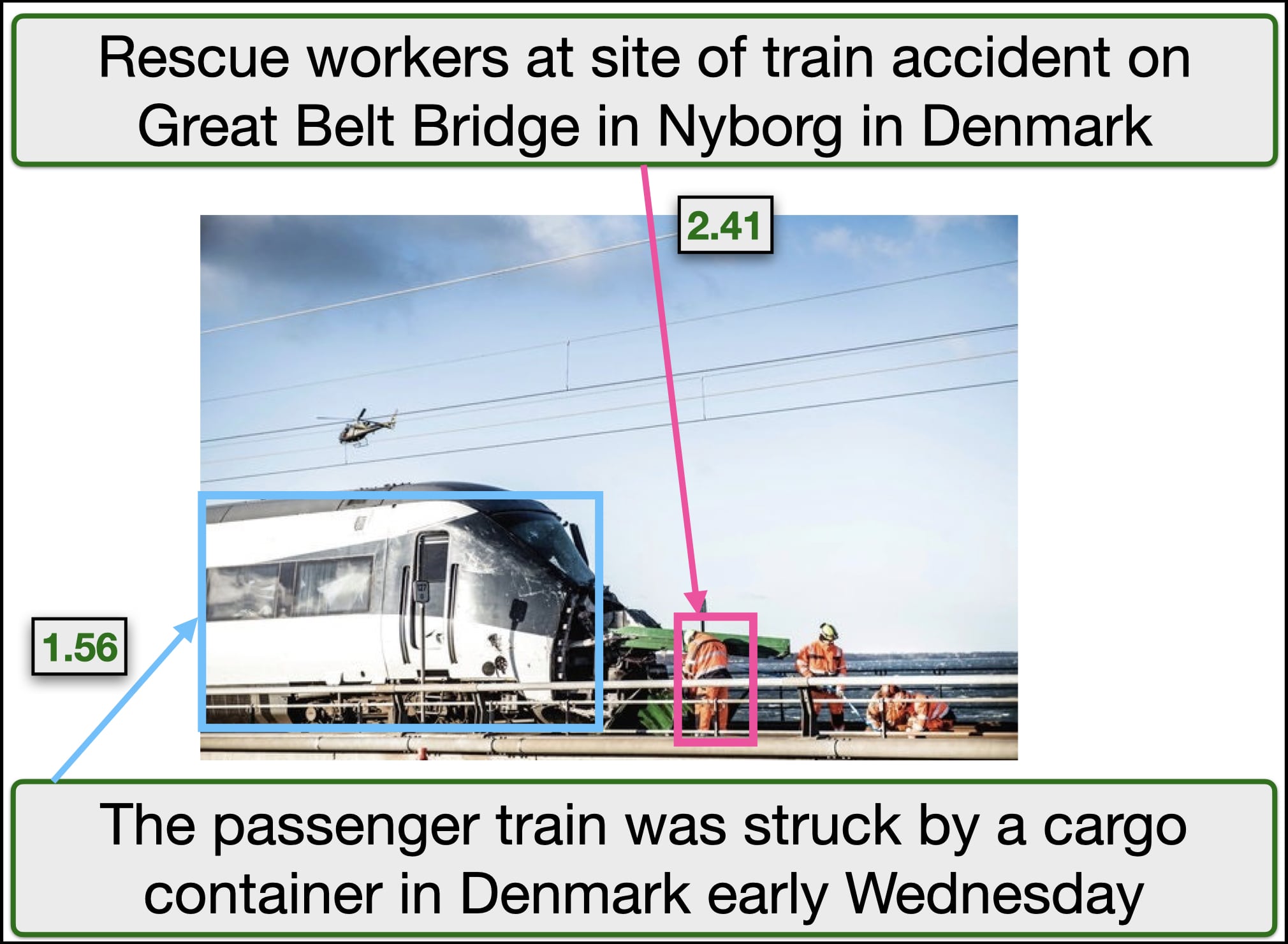} }}%
    \hfill
    \subfloat{{\includegraphics[width=0.32\linewidth]{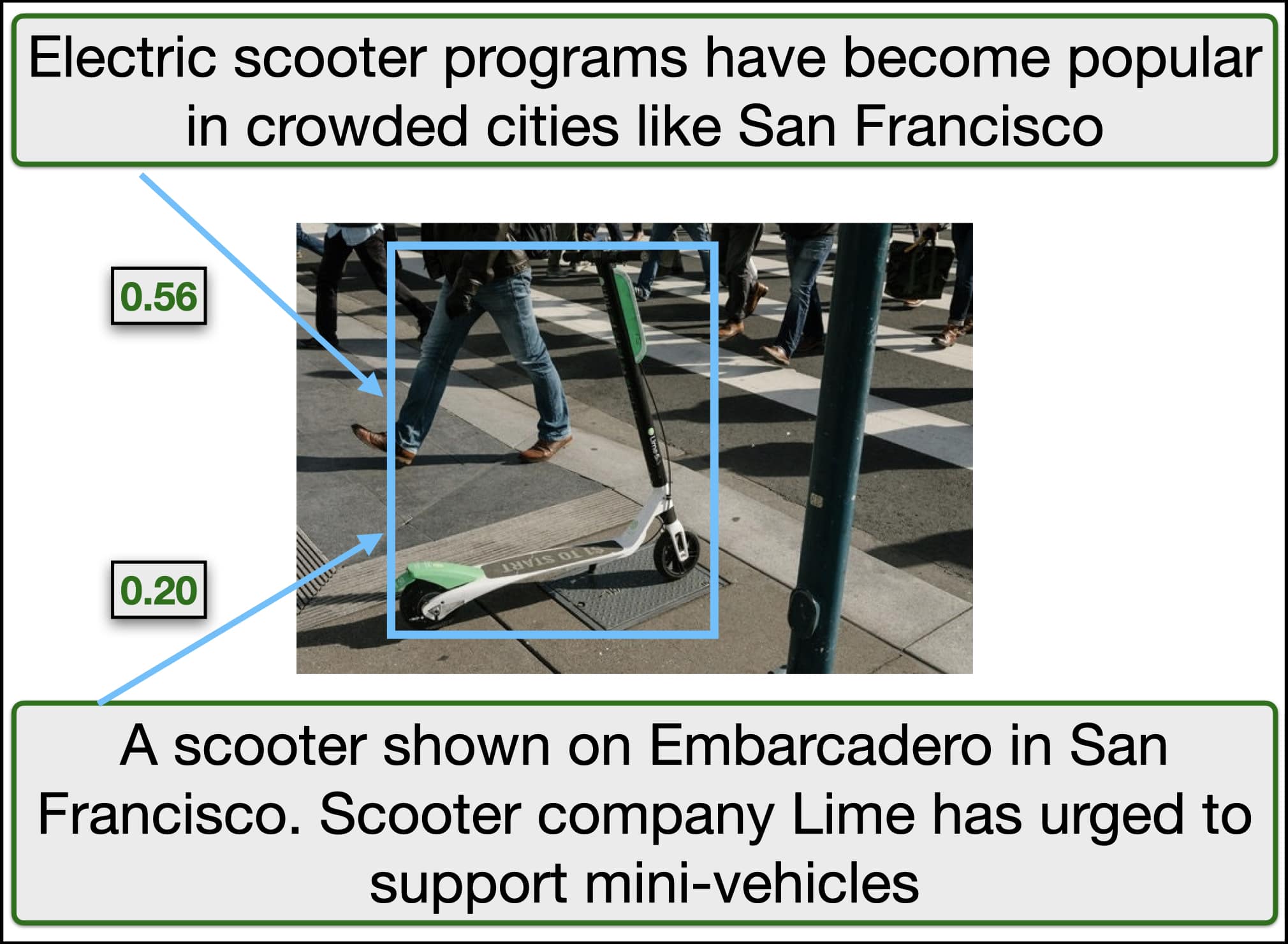} }}%
    \hfill
\end{center}
    \vspace{-0.4cm}
   \caption{Qualitative results of visual grounding of captions with the objects in the image. The top two rows show the grounding for out-of-context pairs and the bottom two rows show the grounding for pairs which are not out of context. We show object-caption scores for two  captions per image. The captions with \greencmd{green} border show the true captions and the captions with \redcmd{red} border show the false caption. Scores indicate association of the most relevant object in the image with the caption.
   }
\label{fig:qualitative_res}
\end{figure*}

\vspace{-0.5cm}
\paragraph{Qualitative Results.}
We visualize grounding scores in Fig.~\ref{fig:qualitative_res} from applying our image-text matching model to several image-caption pairs from the test set. 
The results indicate that our self-supervised matching strategy learns sufficient alignment between objects and captions to perform out-of-context image detection.

\subsection{Out-of-Context Evaluation}\label{sec:ours}

\noindent
\textbf{Which is the best Text Embedding?} 
To evaluate the effect of different text embeddings, we experiment with:
(1) Pre-trained word embeddings including Glove~\cite{pennington2014glove} and FastText~\cite{Bojanowski2017EnrichingWV} embedded via a one-layer LSTM model and (2) the Transformer based Sentence embeddings proposed by USE~\cite{cer-etal-2018-universal}.
The results in Tab.~\ref{tab:sent_embed} show that even though the match accuracy for all the methods is roughly the same (72\%), using sentence embeddings significantly boosts our final out-of-context image detection accuracy of the model by 9\% (from 76\% to 85\%). In addition, we also compare our results with state-of-the-art pretrained language baseline S-BERT~\cite{sbert} and outperform it by a margin of 8\%.

\vspace{-0.2cm}
\begin{table}[ht]
\begin{center}
\begin{tabular}{c|c|c}
\hline
\toprule
Text Embed & Match Acc. & Context Acc. \\
\toprule
S-Bert~\cite{sbert} & - & 0.77 \\
\toprule
Glove~\cite{pennington2014glove} & 0.72 & 0.76 \\
FastText~\cite{Bojanowski2017EnrichingWV} & 0.71 & 0.78  \\
USE~\cite{cer-etal-2018-universal} & 0.72 & \textbf{0.85}  \\
\toprule
\end{tabular}
\end{center}
\vspace{-0.4cm}
\caption{Ablation with different text embeddings. Top row shows pre-trained S-Bert~\cite{sbert} language baseline evaluated on our test set.}%
\label{tab:sent_embed}
\end{table}

\noindent
\vspace{-0.05cm}
\textbf{How much training data is needed?} 
Next, we analyze the effect of the available training data corpus for self-supervised training. 
We experiment with different percentages of training data size (w.r.t. to our full data) in Tab.~\ref{tab:dataset_size}.
For these experiments, we use a pretrained USE~\cite{cer-etal-2018-universal} embedding with a one-layer FC model to encode text.
We observed that a larger training set significantly improves the performance of the model. 
For instance, training with full-dataset (160k images) improves \ooc\ detection accuracy by an absolute 13\% (from 72\% to 85\%) compared to a model trained only with 10\% of the dataset (16K images), thus benefiting from the large diversity in the dataset.
We also notice that a relatively higher match accuracy leads to better \ooc\ detection accuracy. 
This suggests that our proposed self-supervised learning strategy (trained with match vs no-match loss) effectively helps to improve \ooc\ detection accuracy.

\vspace{-0.2cm}
\begin{table}[ht]
\begin{center}
\begin{tabular}{l|c|c|c}
\toprule
Dataset  & Train Images & Match  & {Context Acc.} \\
Size & & Acc. & \\
\toprule
Ours (10\%) & 16K & 0.64  &  0.72  \\
Ours (20\%) & 32K & 0.65  &  0.74  \\
Ours (50\%) & 80K & 0.68  & 0.77 \\
Ours (100\%) & 160K & 0.72 & \textbf{0.85}  \\
\toprule
\end{tabular}
\end{center}
\vspace{-0.5cm}
\caption{Ablation with variations in number of train images w.r.t. our full corpus of \COUNTIMAGES{} (160K train, 40k val) images. Using all available data achieves the best results.} %
\label{tab:dataset_size}
\end{table}

\paragraph{Comparison with alternative approaches.}
Finally, we compare our best-performing model with other baselines, in particular, methods that work on rumor detection.
Most other fake news detection methods are supervised,
where the model takes an image and a caption as input and predicts the class label. 
EANN~\cite{Wang_eann_18} and Jin et al.~\cite{zhiwei_17} were proposed specifically for Rumor/Fake News Classification; however, EmbraceNet~\cite{choi2019embracenet} is a generic multi-modal classification method. 
Since neither of these methods perform self-supervised out-of-context image detection using object features (using bounding boxes), an out-of-the-box comparison is not feasible, and we must adapt these methods for our task.
Following our training setup (Sec.~\ref{sec:method_train}), we first train these models for the binary task of image-text matching with the network architecture and losses proposed in their original papers. 
During test time, we then use GradCAM~\cite{SelvarajuCDVPB17} to construct bounding boxes around activated image regions and perform out-of-context detection as described in Sec.~\ref{sec:method_test}. 
The results in Tab.~\ref{tab:tab3} show that our model outperforms previous fake news detection methods for out-of-context detection by a large margin of 14\% (from 71\% to 85\%). 
Overall, we achieve up to 85\% \ooc\ image detection accuracy.

\begin{table}[ht]
\begin{center}
\begin{tabular}{c|c|c}
\hline
\toprule
Method & Match Acc. & {Context Acc.} \\
\toprule
EANN~\cite{Wang_eann_18} & 0.57  & 0.63    \\
EmbraceNet~\cite{choi2019embracenet} & 0.59  & 0.68  \\
Jin \etal ~\cite{zhiwei_17} & 0.60  & 0.71   \\
Ours & 0.72 & \textbf{0.85}   \\
\toprule
\end{tabular}
\end{center}
\vspace{-0.5cm}
\caption{We compare our method against three state-of-the-art methods. Our method outperforms all other methods, resulting in 85\% out-of-context detection accuracy.}
\label{tab:tab3}
\end{table}
\vspace{-0.4cm}
\section{Conclusions}
We have introduced an automated method to detect \ooc\ images with respect to textual descriptions.
Our key insight is to ground text to the image, as language-only analysis cannot effectively interpret semantically different captions that do not conflict due to referring to different objects in the image.
Our approach thus ties two potential captions for an image to corresponding  object regions for \ooc\ determination, reaching up to 85\% detection accuracy.
We adopt a self-supervised training strategy to learn strong localization features based only on a set of captioned images, without the need for explicit out-of-context annotations.
We further introduce a new dataset to benchmark this \ooc\ task.
Overall, we believe that our method takes an important step towards addressing misinformation in online news and social media platforms, thus supporting and scaling up fact-checking work.
In particular, we hope that our new dataset, which we will publish along with this work, will lay a foundation to continue research along these lines to help online journalism and improve social media.
%


\section*{Acknowledgments}
{
This work is supported by a TUM-IAS Rudolf M\"o{\ss}bauer Fellowship and the ERC Starting Grant Scan2CAD (804724). We would also like to thank New York Times Team for providing API support for dataset collection, Google Cloud for GCP credits, Farhan Abid for dataset annotation, and Angela Dai for video voice over. 
}

{\small
\bibliographystyle{ieee_fullname}
\bibliography{egbib}
}

\newpage
\appendix

\section*{Supplemental Material}

In this supplemental document, we include additional details regarding the terminology used in our main paper; see Section~\ref{sec:def}. 
In addition, in Section~\ref{sec:dataset_details}, we document details regarding dataset cleanup/annotation and dataset statistics. We also visualize image-caption groundings for additional images from the test set. We then provide experimental details and metrics used for evaluation in our main paper in Section~\ref{sec:setup}. Finally, in Section~\ref{sec:extra_experiments}, we perform ablations with different augmentations on our dataset and experiments on RefCOCO~\cite{yu2016modeling} to evaluate which configuration is most effective to learn better object-caption groundings.

\section{Definitions}\label{sec:def}

\begin{itemize}
\item {\em Caption:} 
The fact-checking community uses multiple terms for the accompanying text that describes what is in the image, including the term ``caption'' or ``claim''.
In this paper, we only use the word {\em caption} for the textual image descriptions.  We do not use the word {\em claim}.
\item {\em \Cic}: 
Describing falsely what is supported by an image with the aim to mislead audiences is often labelled by the fact-checkers as ``misleading context'', ``out-of-context'', ``transformed context'', ``context re-targeting'', ``re-contextualization'', ``misrepresentation'', and possibly other terms.  
A news article or social media post that is in question only shows the image with one caption and some fact-checking articles also only show a single instance of a false caption, but most fact-checks show at least two captions, usually one correct caption from the original dissemination of the image, and also the false caption.  
We describe an image to be {\em \ooc\ } relative to two contradicting captions and we describe these two captions as {\em \cic}.  
However, we do not aim to determine which on of the two captions is false or true, or possibly if both captions are false.  This is up to future research.
\end{itemize}

\begin{figure*}[tbh!]
\begin{center}
\includegraphics[width=1.0\textwidth]{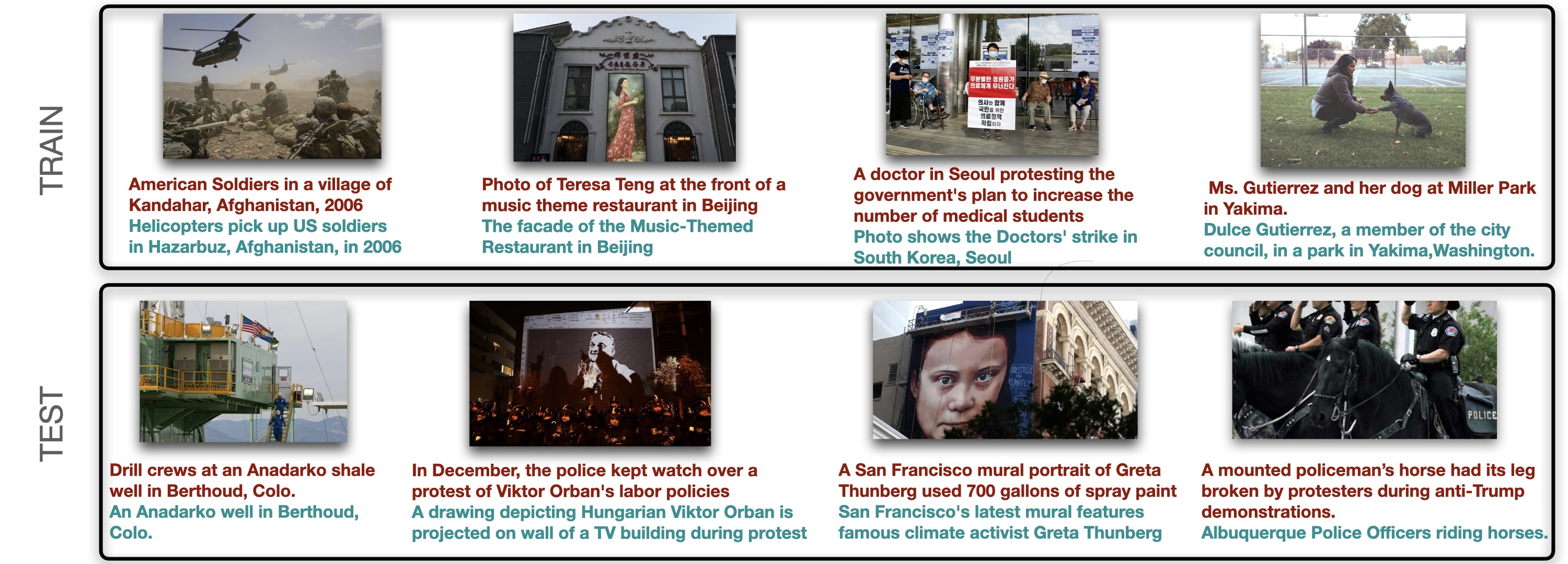}
\end{center}
\vspace{-0.4cm}
   \caption{Dataset: images and captions from the train set (top row); samples from the test set (bottom row). }
\label{fig:dataset_samples}
\end{figure*}

\begin{figure*}[ht]
\begin{center}
\vspace{-0.2cm}
\subfloat{{\includegraphics[width=0.32\linewidth]{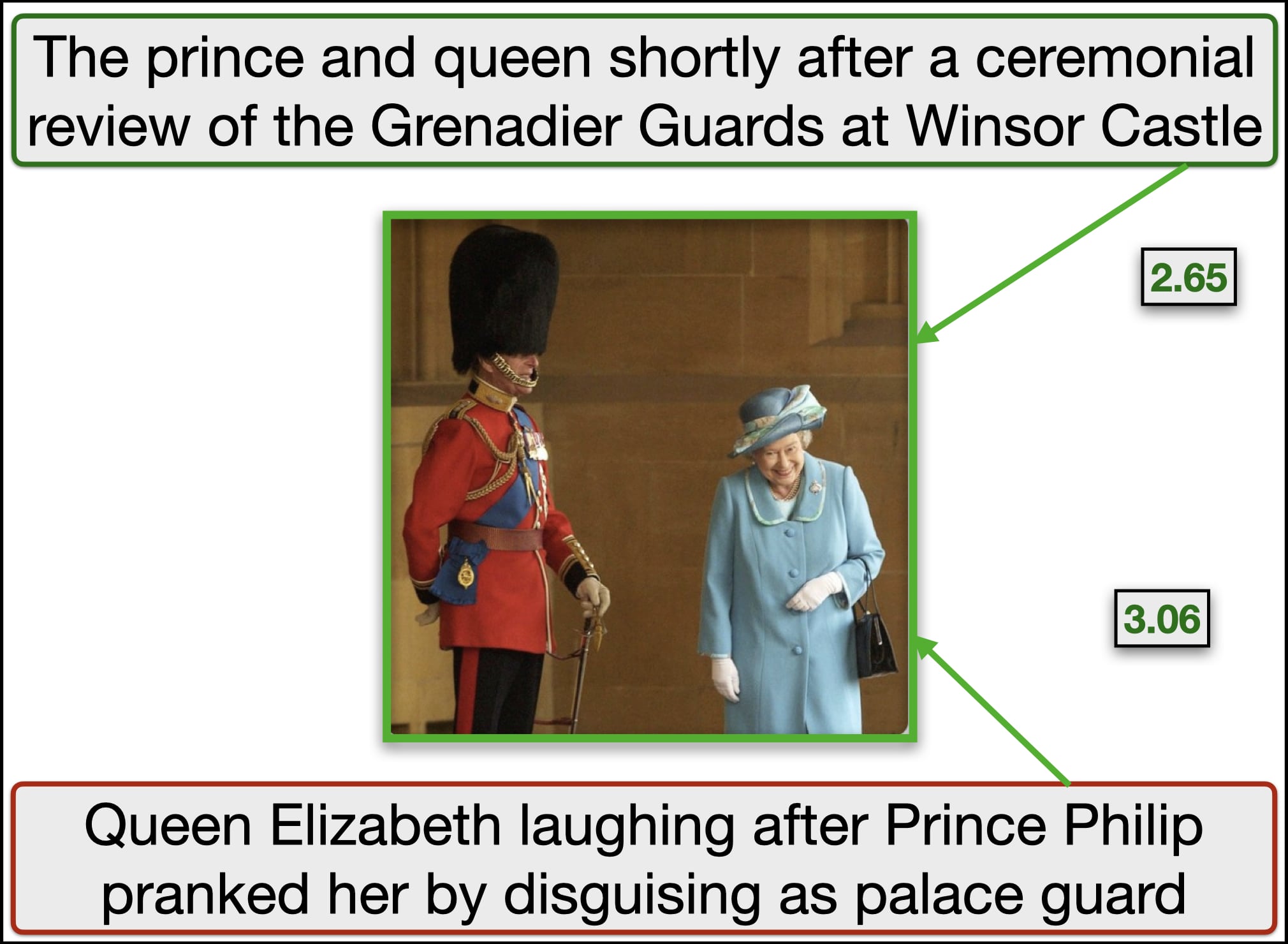} }}%
    \hfill
    \subfloat{{\includegraphics[width=0.32\linewidth]{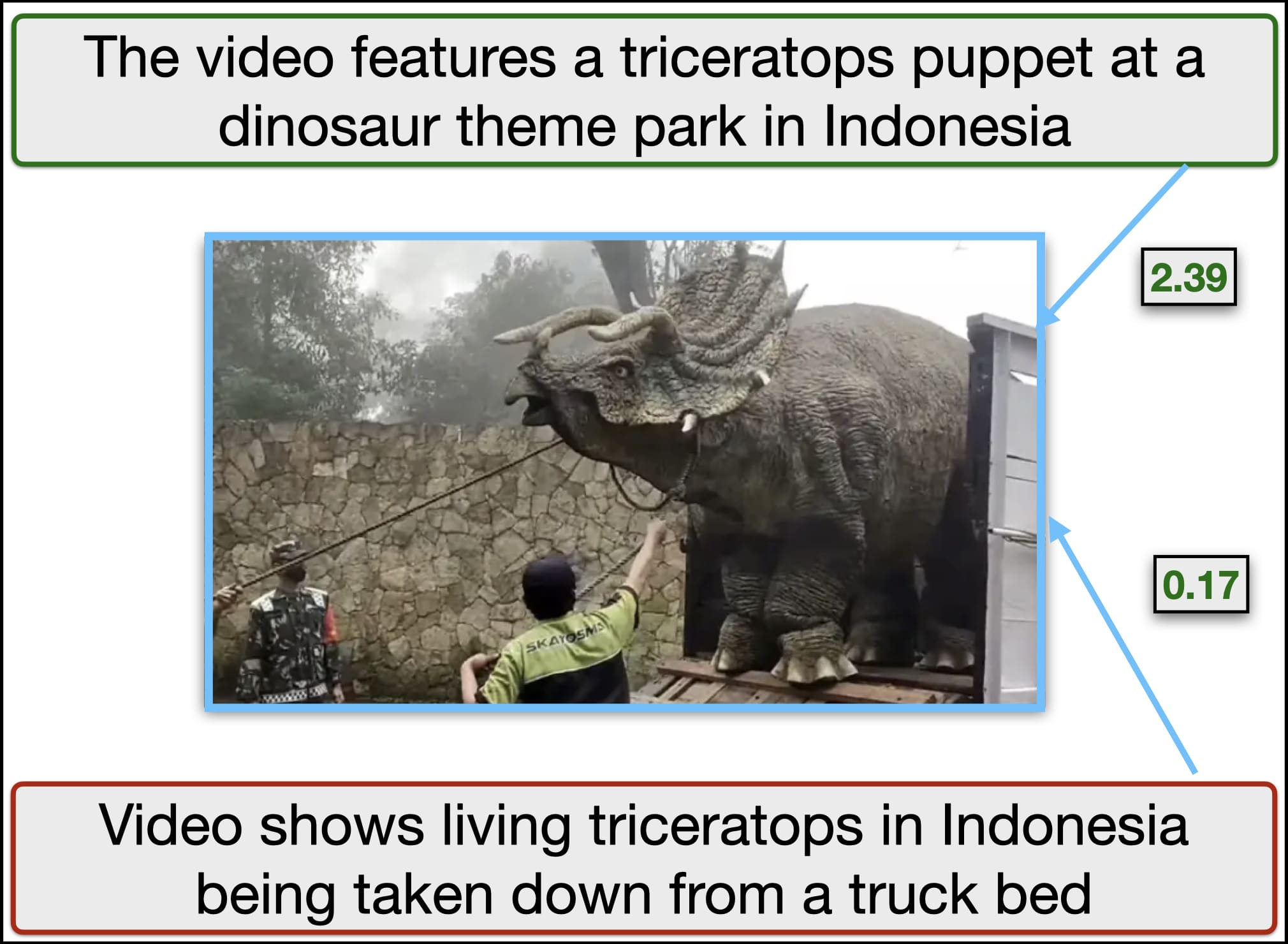} }}%
    \hfill
    \subfloat{{\includegraphics[width=0.32\linewidth]{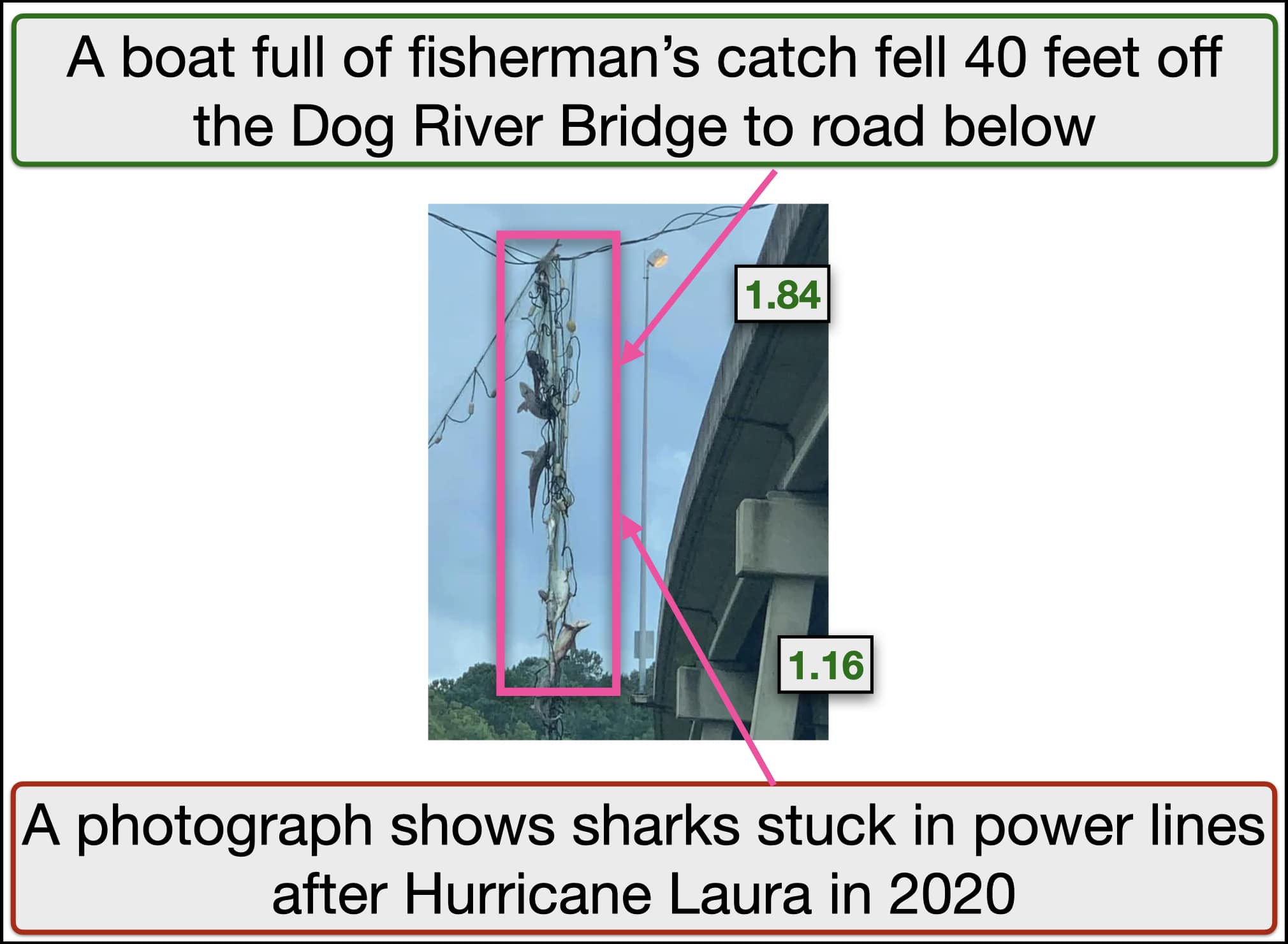} }}
    \end{center}
    
    \begin{center}
    \subfloat{{\includegraphics[width=0.32\linewidth]{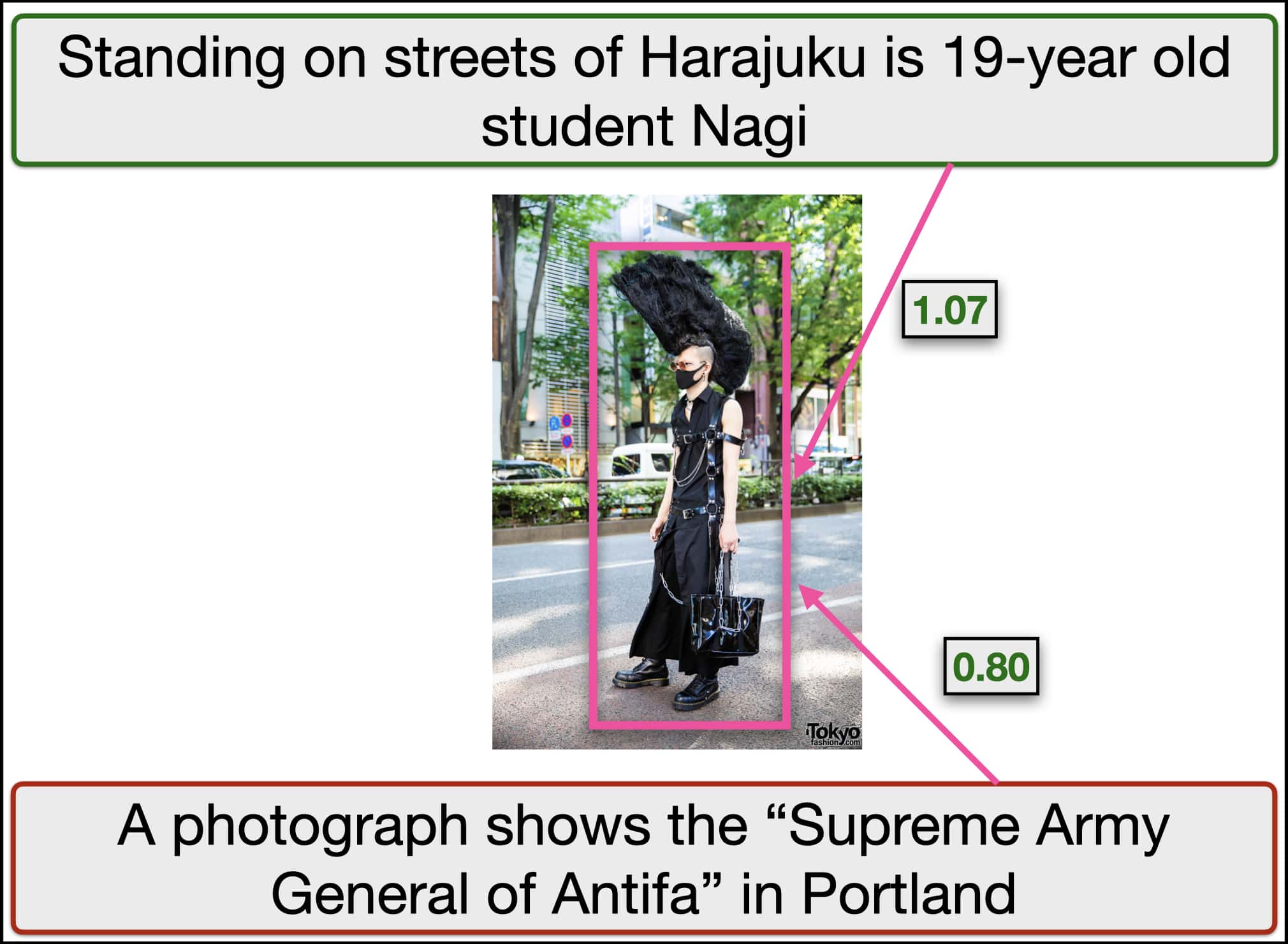} }}%
    \hfill
    \subfloat{{\includegraphics[width=0.32\linewidth]{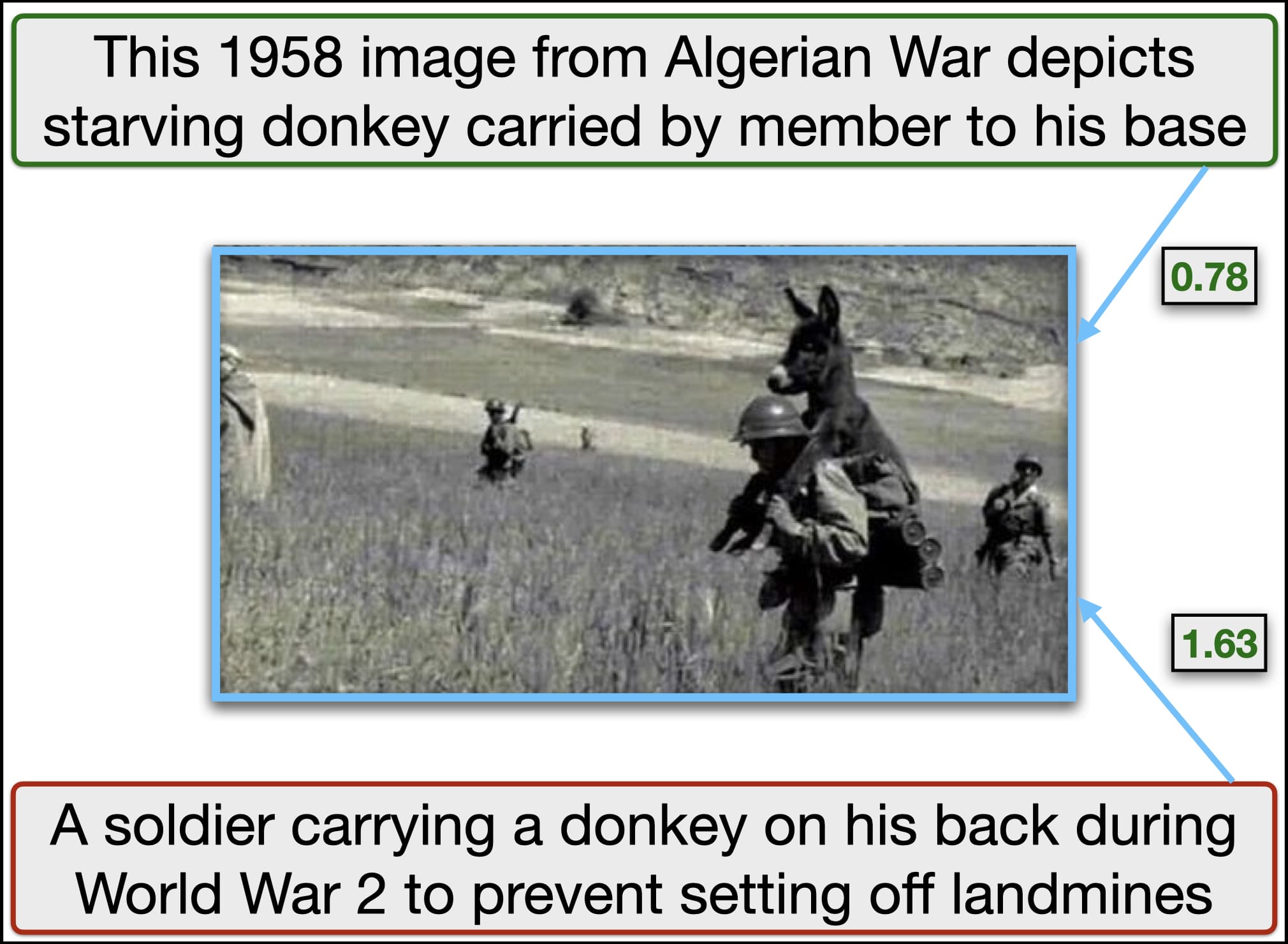} }}%
    \hfill
    \subfloat{{\includegraphics[width=0.32\linewidth]{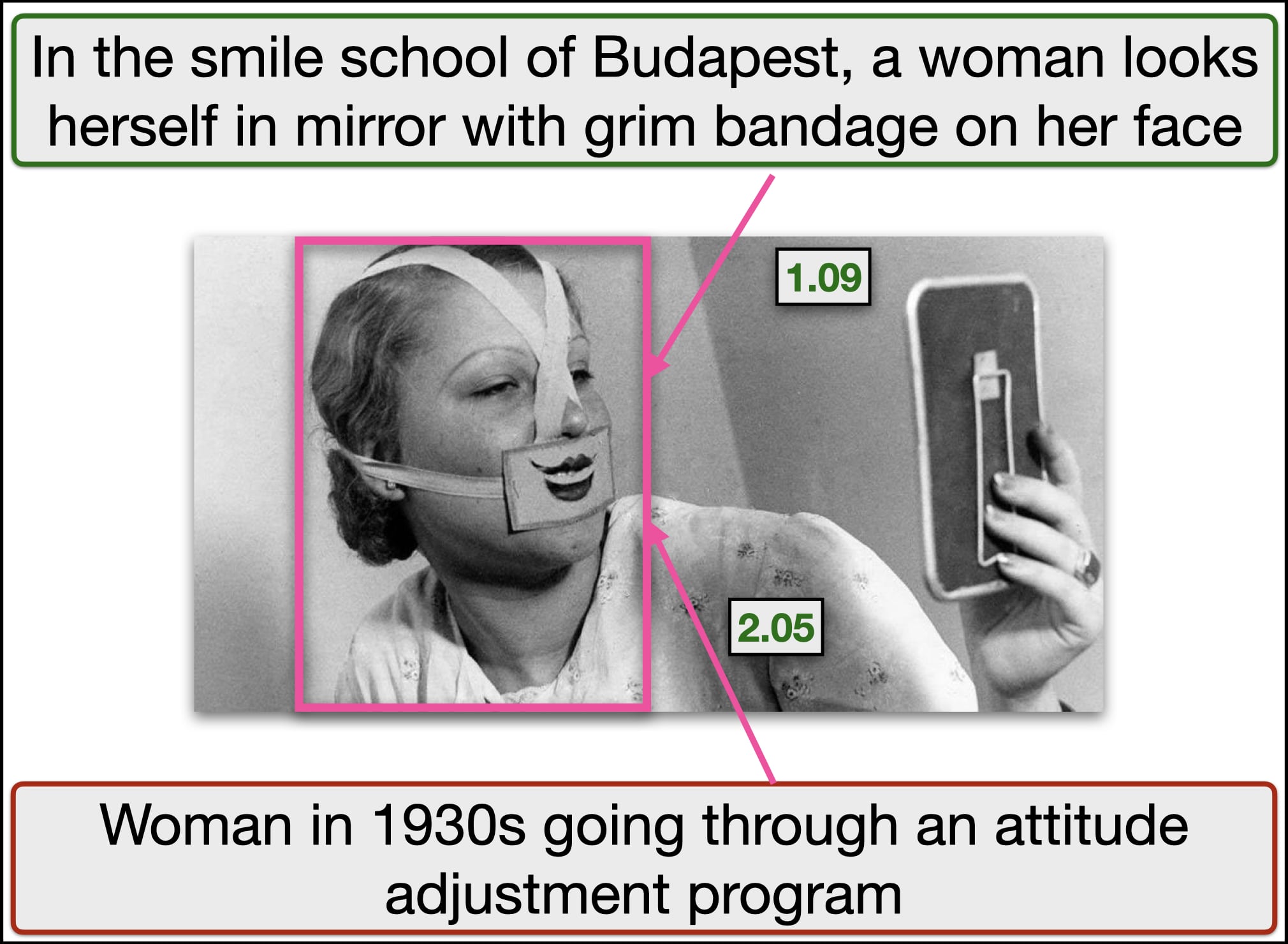} }}%
    \hfill
    \end{center}
    \begin{center}
    \subfloat{{\includegraphics[width=0.32\linewidth]{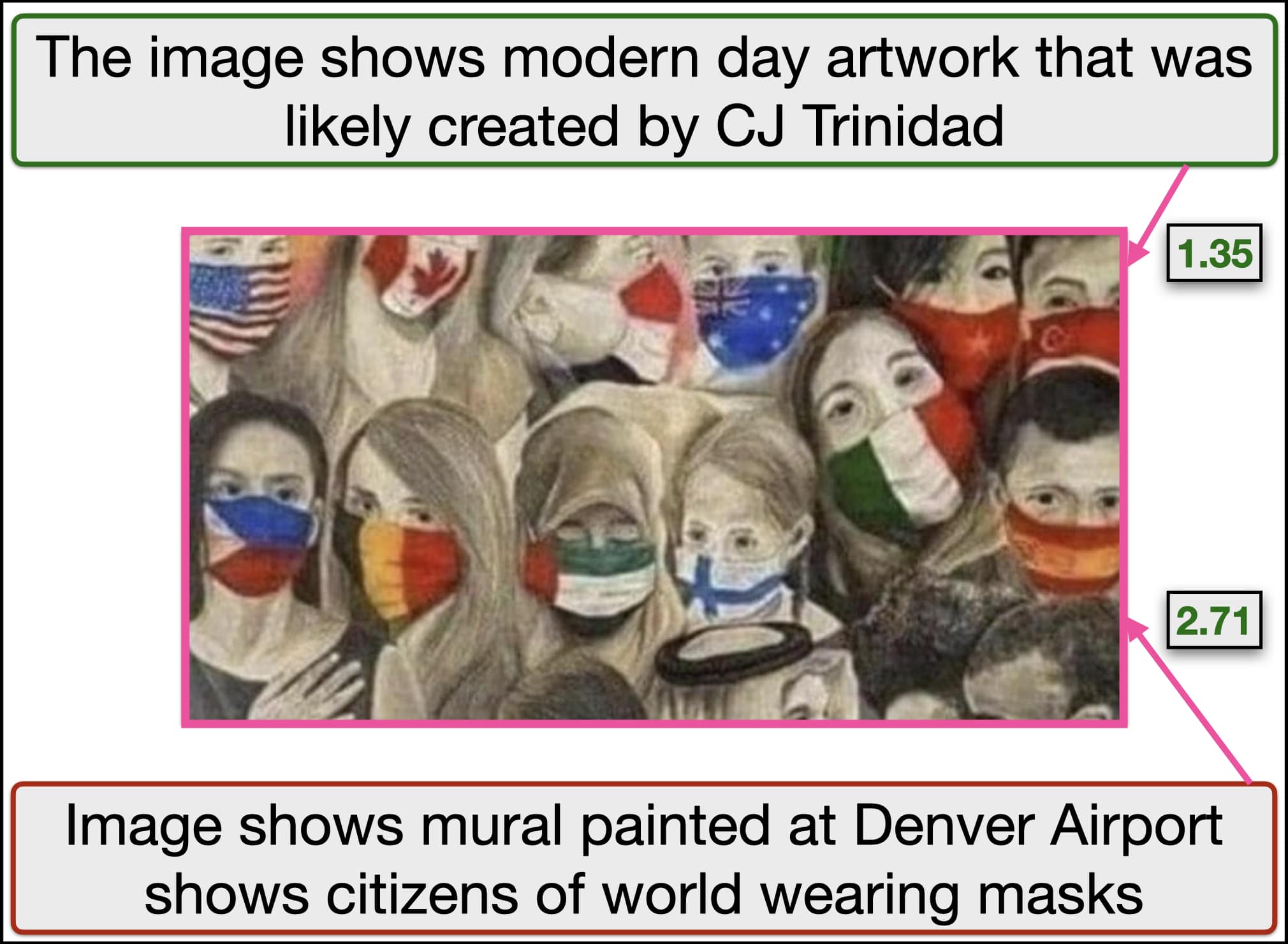} }}%
    \hfill
    \subfloat{{\includegraphics[width=0.32\linewidth]{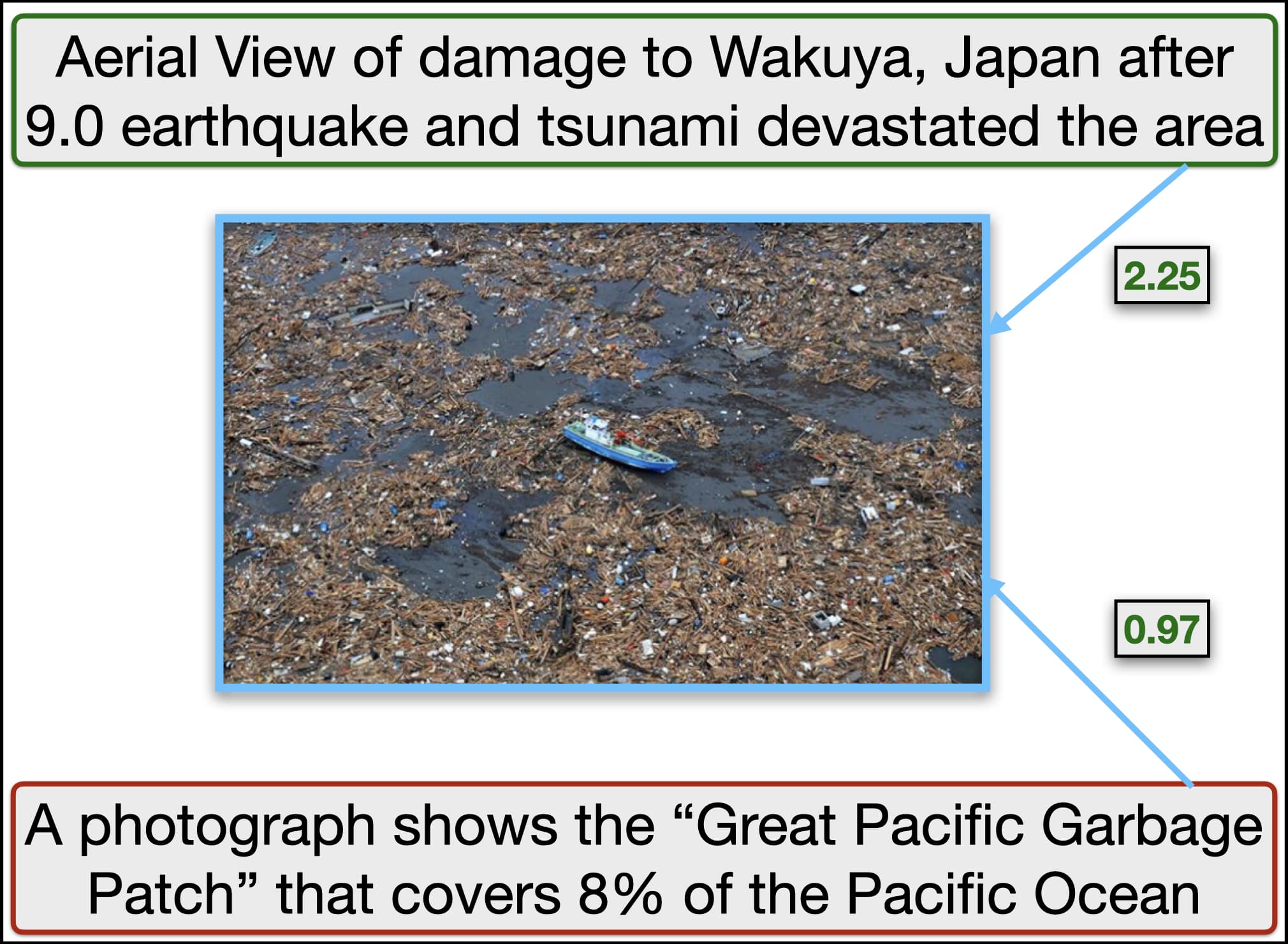} }}%
    \hfill
    \subfloat{{\includegraphics[width=0.32\linewidth]{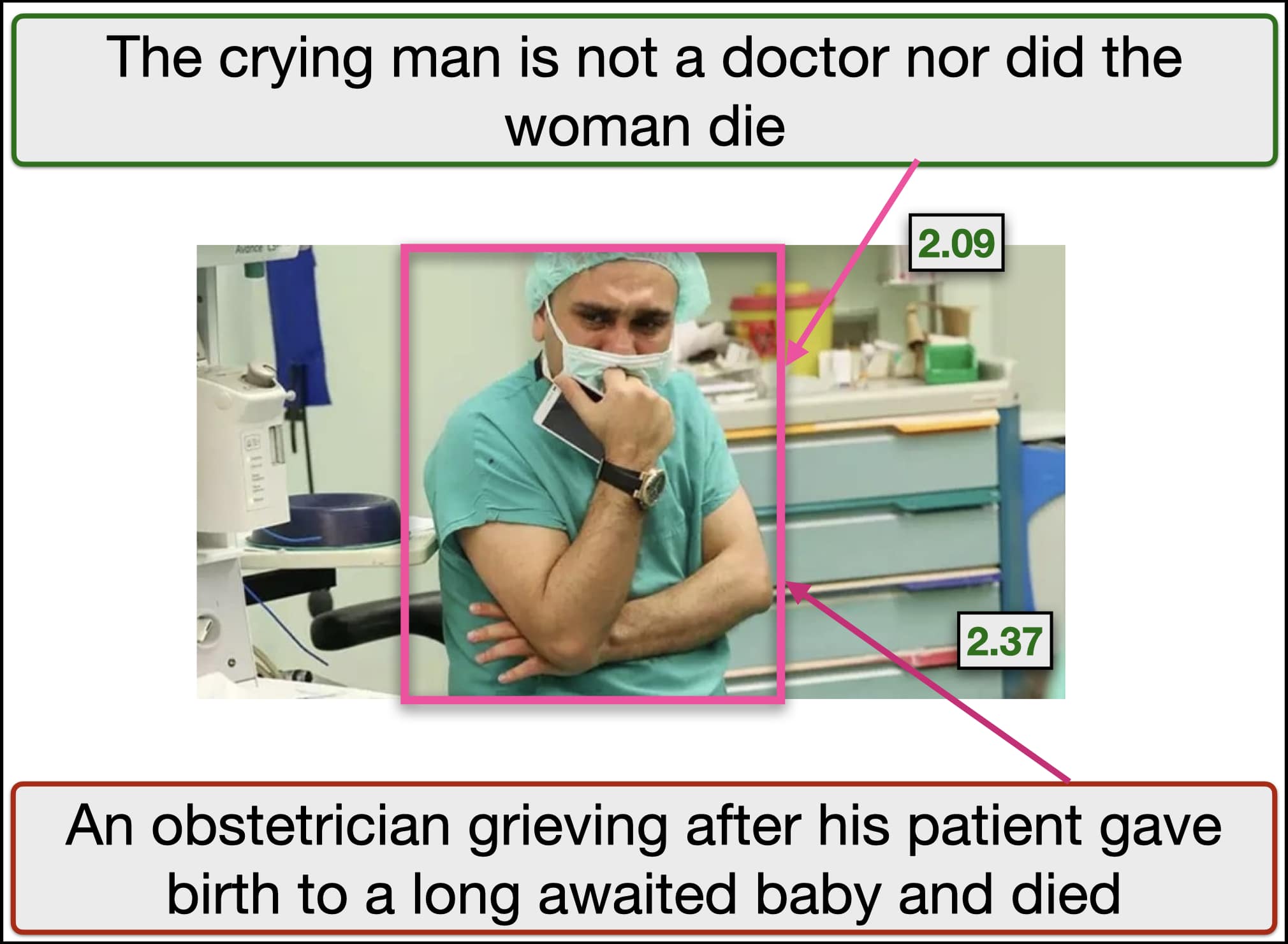} }}%
    \hfill
\end{center}
\begin{center}
    \subfloat{{\includegraphics[width=0.32\linewidth]{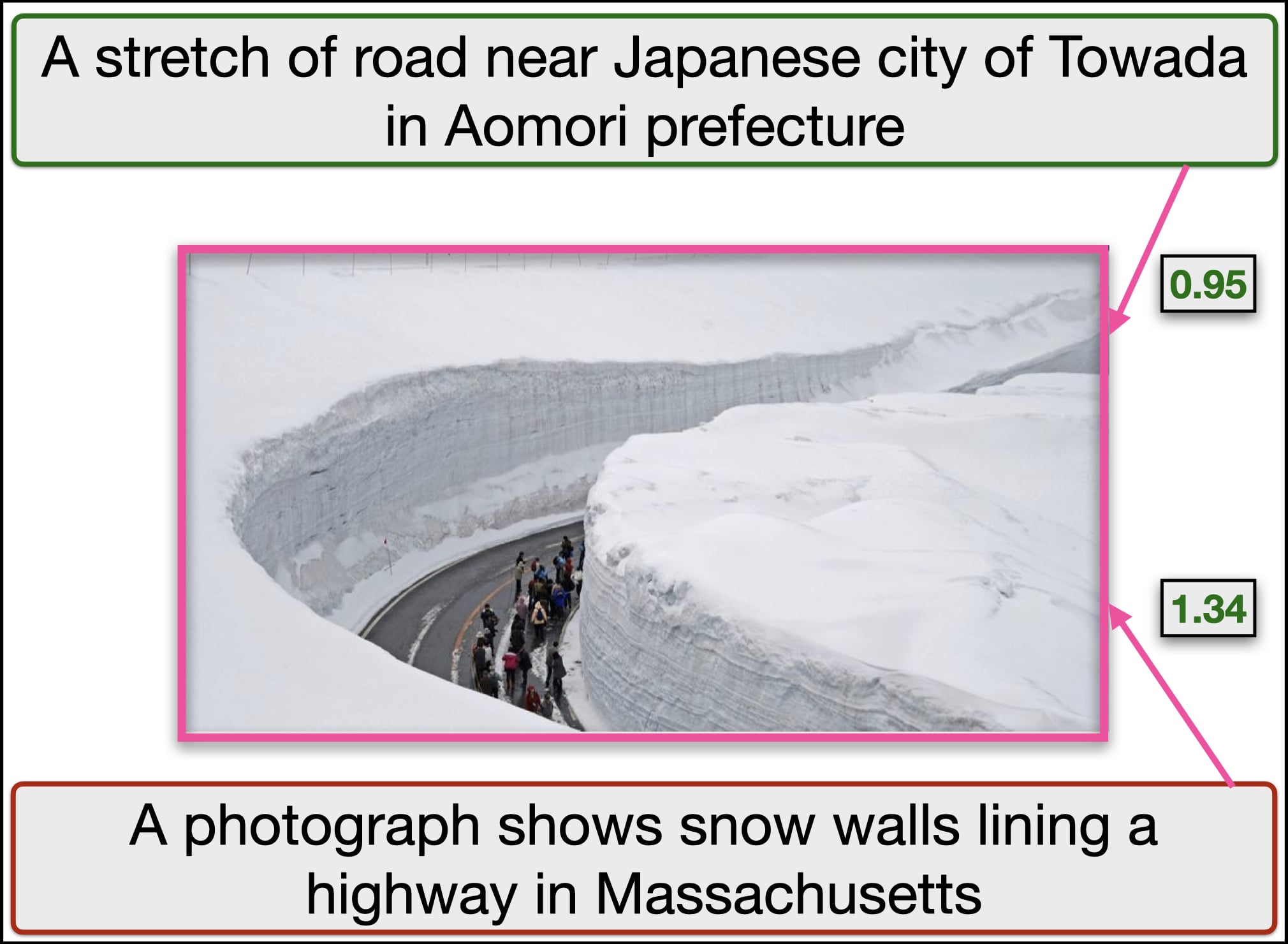} }}%
    \hfill
    \subfloat{{\includegraphics[width=0.32\linewidth]{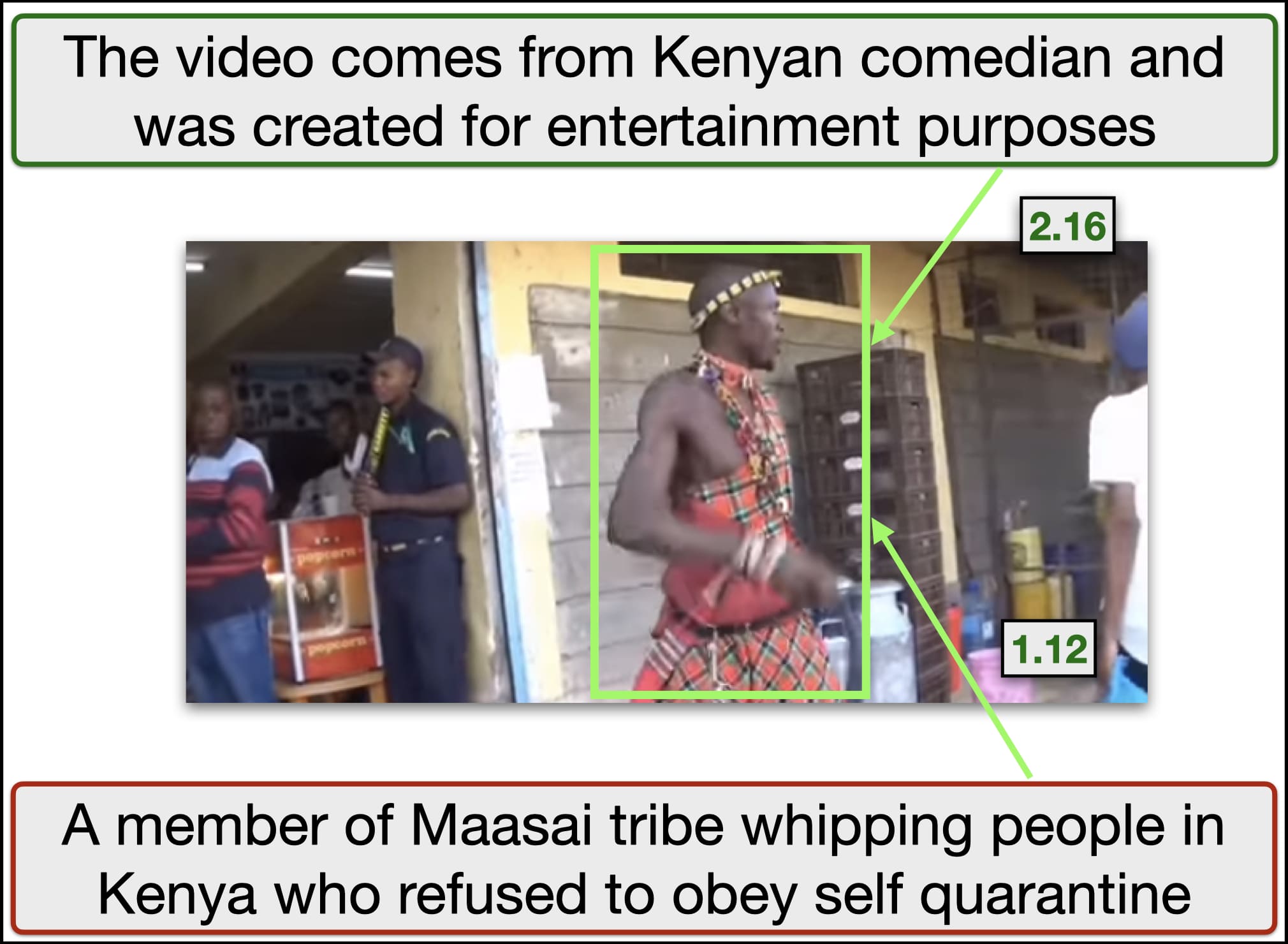} }}%
    \hfill
    \subfloat{{\includegraphics[width=0.32\linewidth]{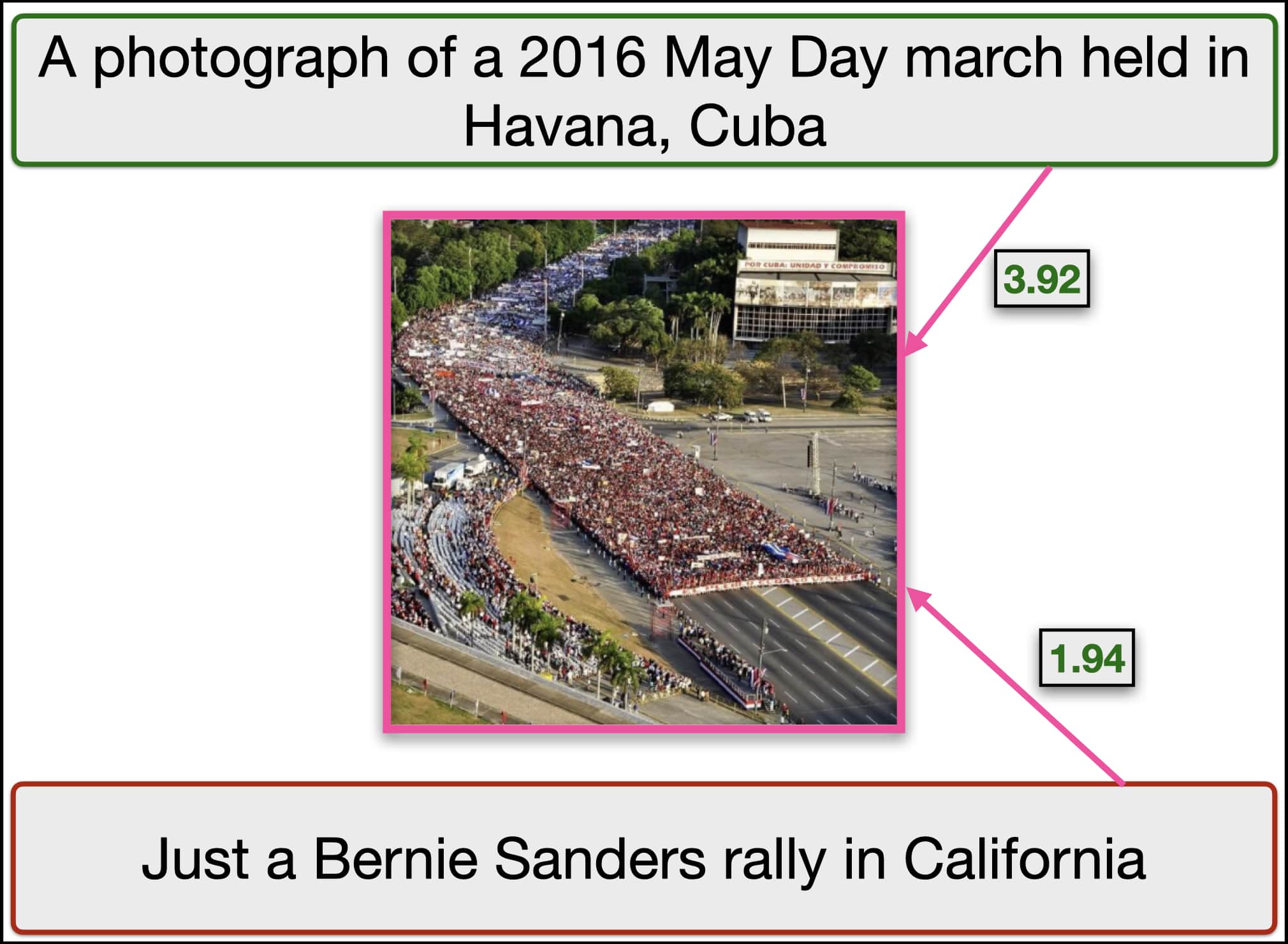} }}%
    \hfill
\end{center}
    \vspace{-0.4cm}
   \caption{Qualitative results of visual grounding of captions with the objects in the image for \Ooc{} image-caption pairs. We show object-caption scores for two captions per image. The captions with \greencmd{green} border show the true captions and captions with \redcmd{red} border show the false/misleading captions. Scores indicate association of the most relevant object in the image with the caption.
   }
\label{fig:qualitative_res_ooc}
\end{figure*}

\begin{figure*}[ht]
\begin{center}
\vspace{-0.2cm}
\subfloat{{\includegraphics[width=0.32\linewidth]{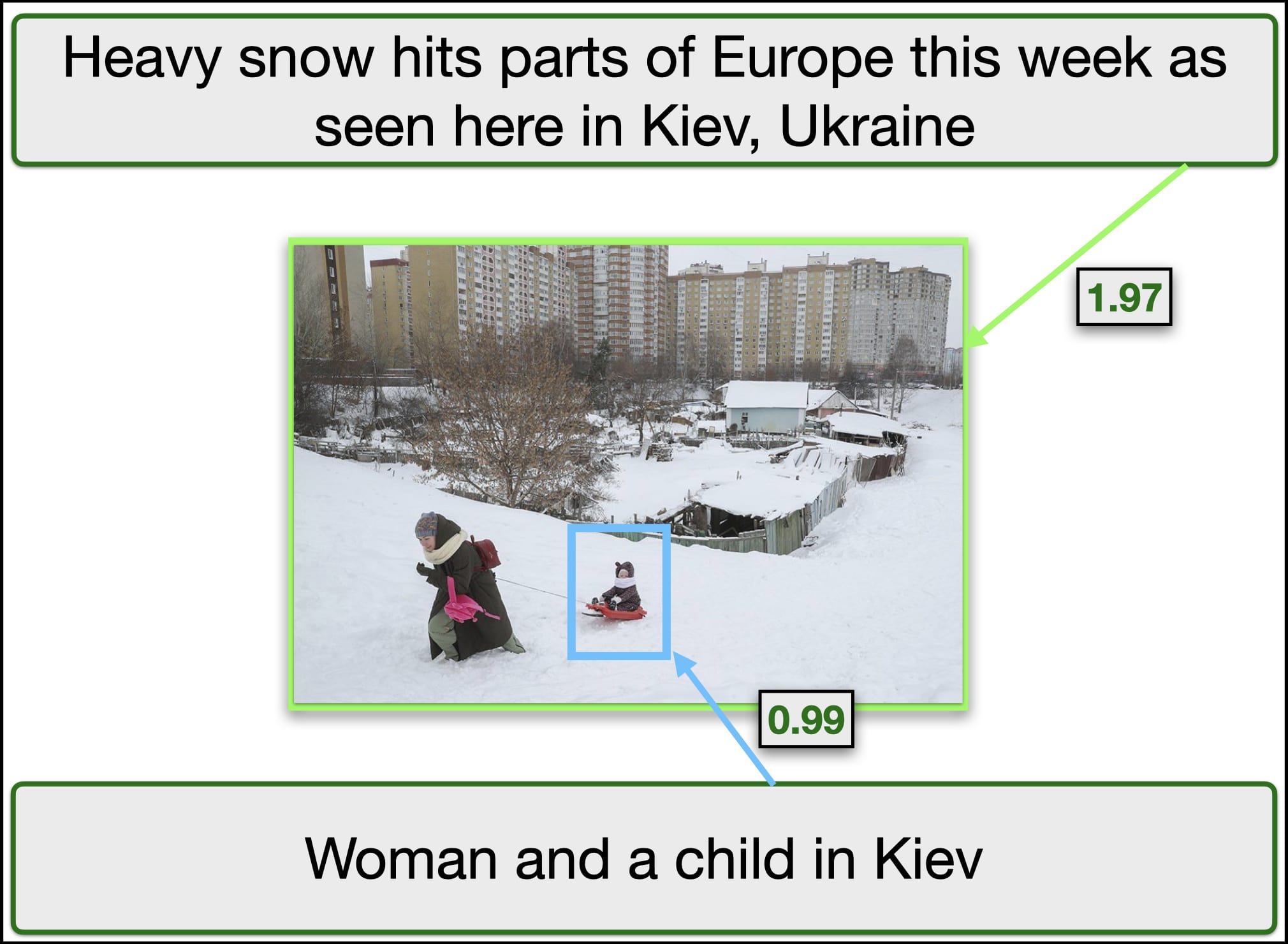} }}%
    \hfill
    \subfloat{{\includegraphics[width=0.32\linewidth]{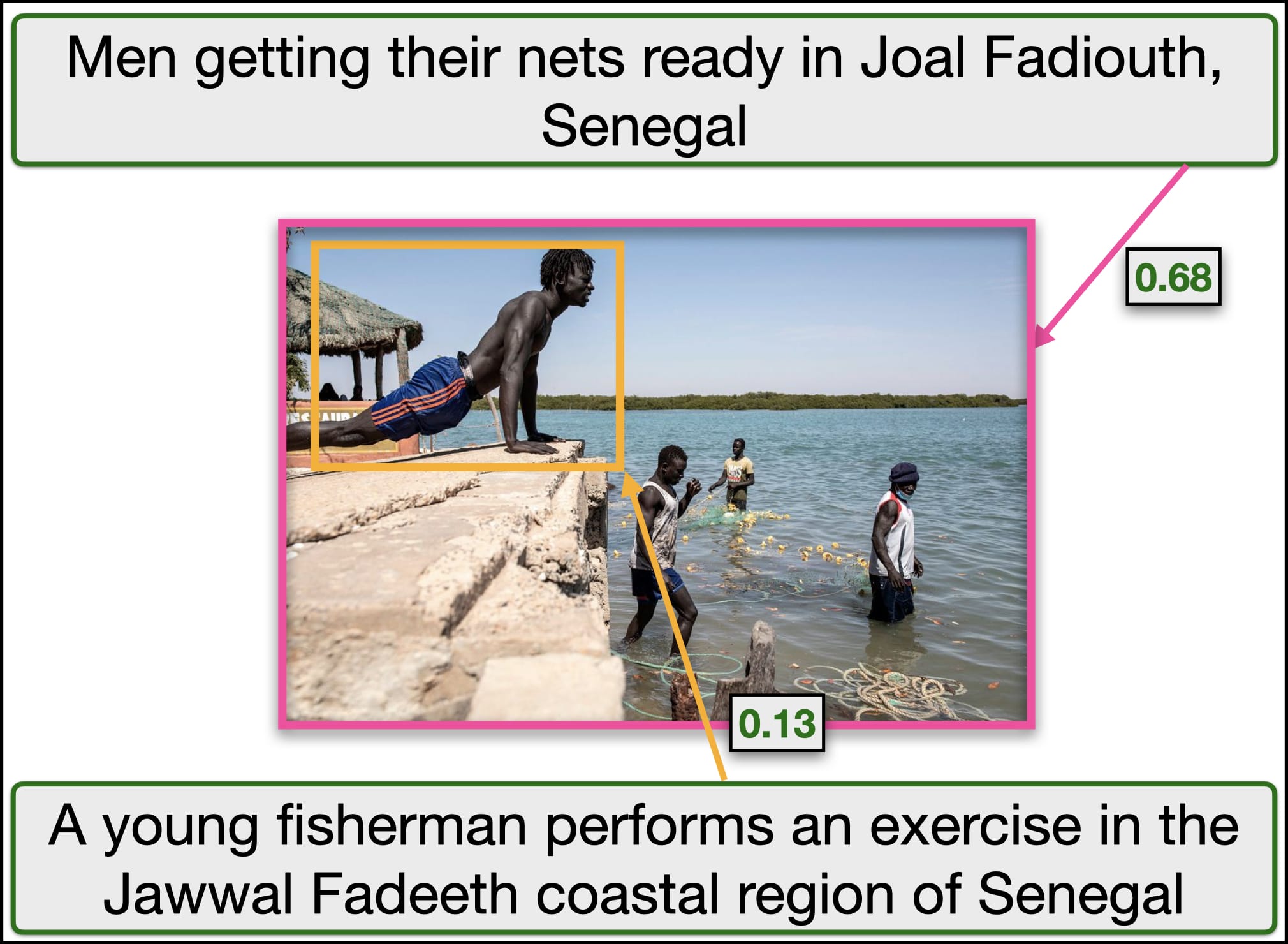} }}%
    \hfill
    \subfloat{{\includegraphics[width=0.32\linewidth]{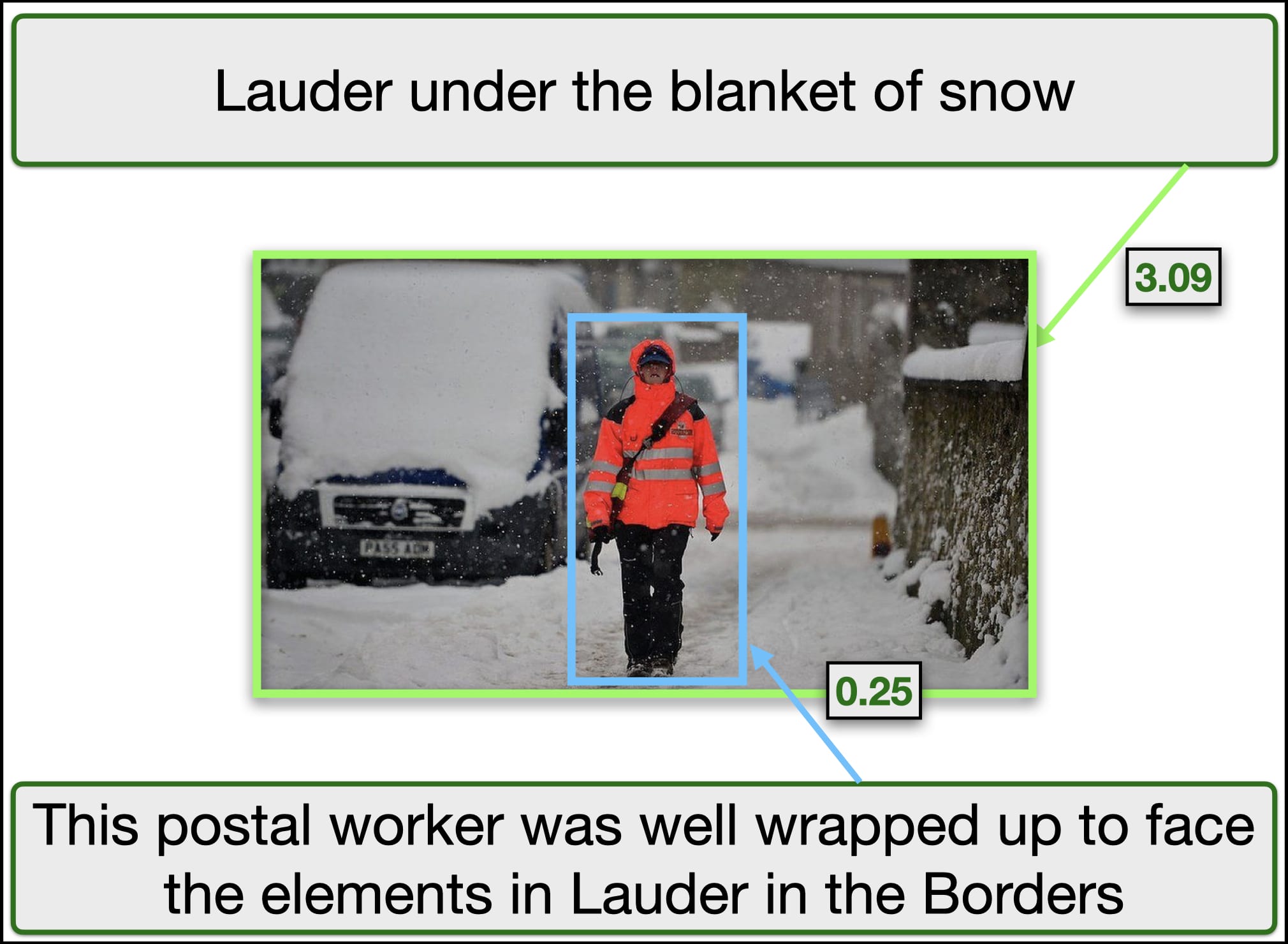} }}
    \end{center}
    
    \begin{center}
    \subfloat{{\includegraphics[width=0.32\linewidth]{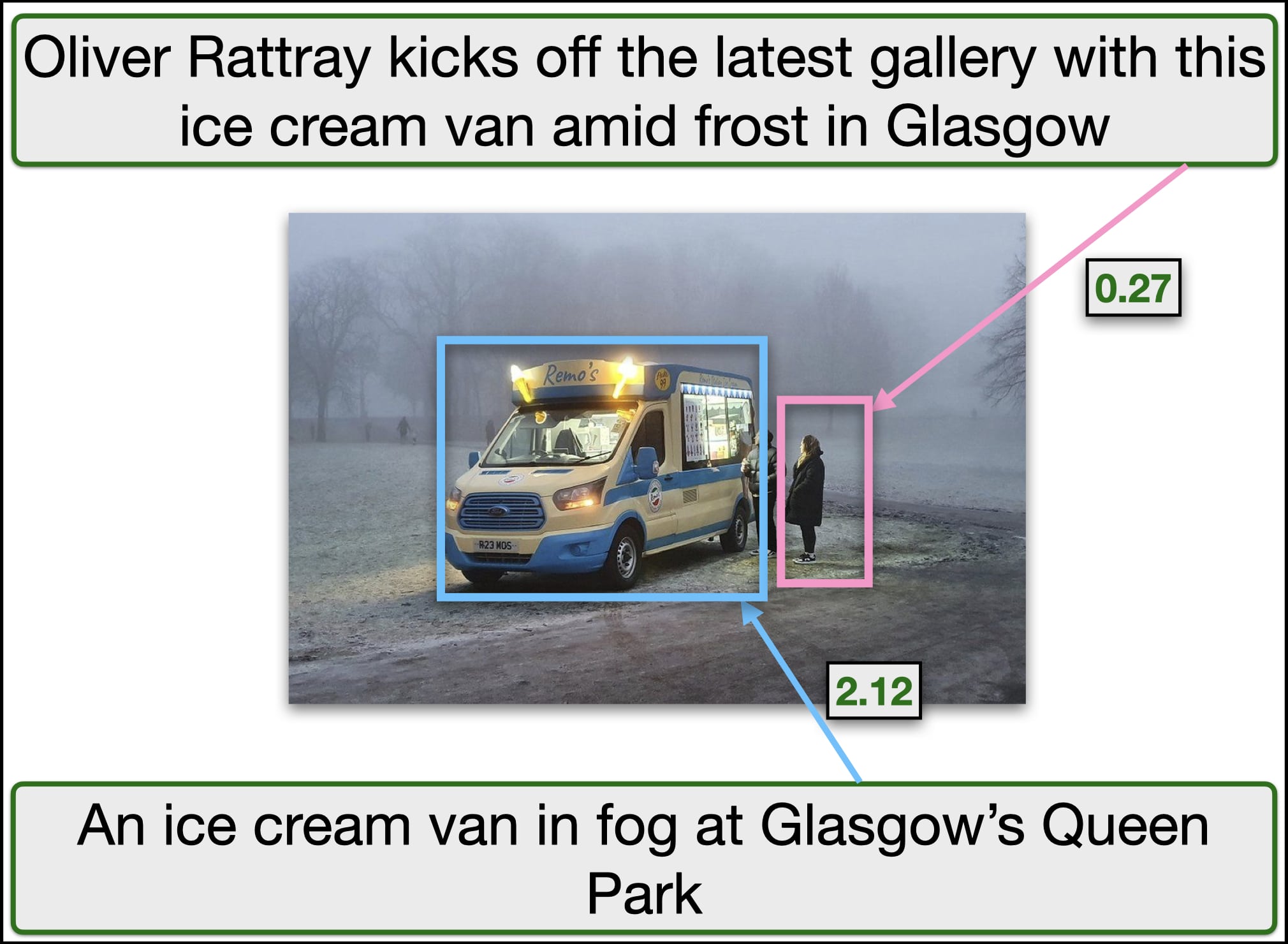} }}%
    \hfill
    \subfloat{{\includegraphics[width=0.32\linewidth]{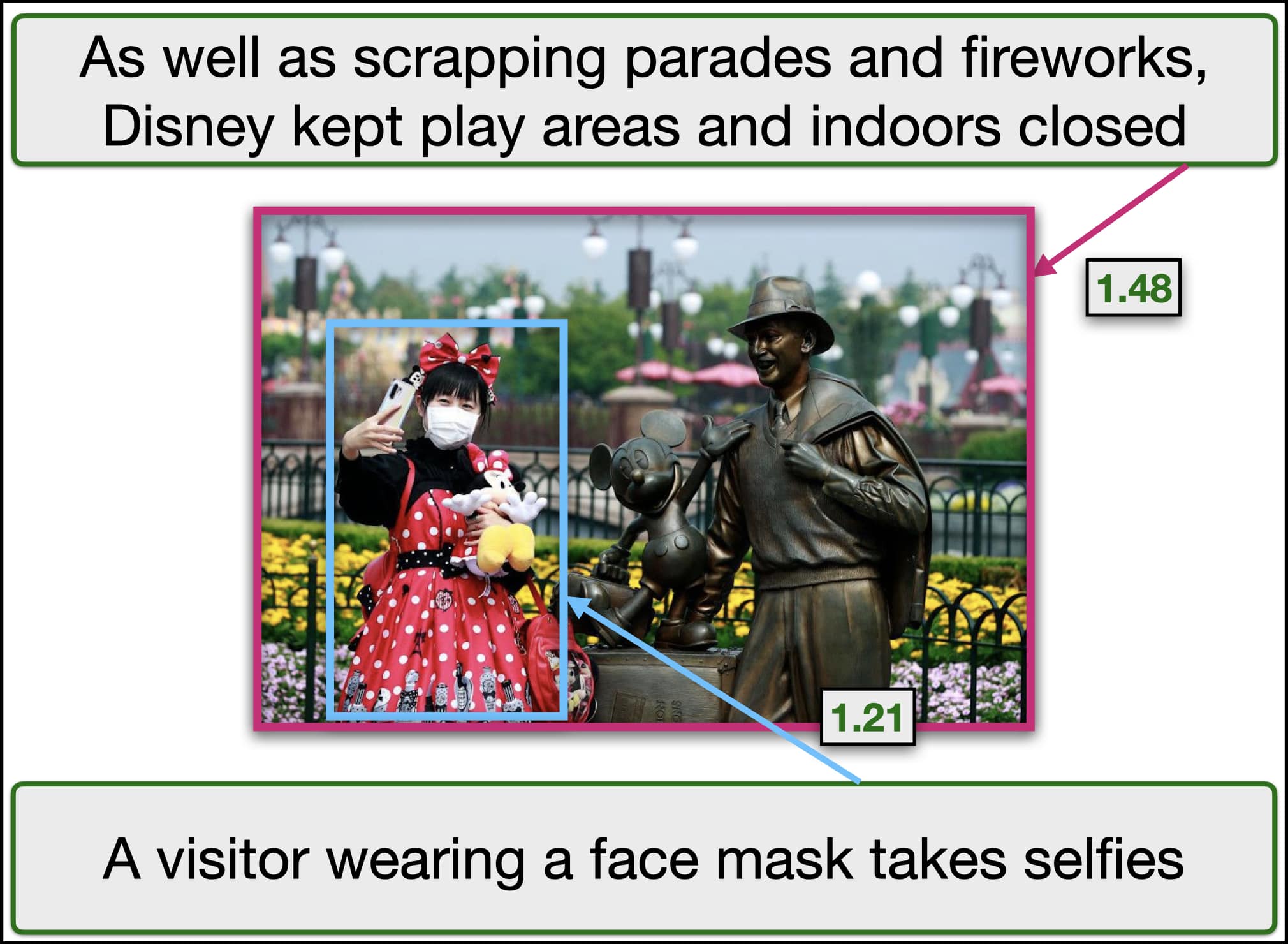} }}%
    \hfill
    \subfloat{{\includegraphics[width=0.32\linewidth]{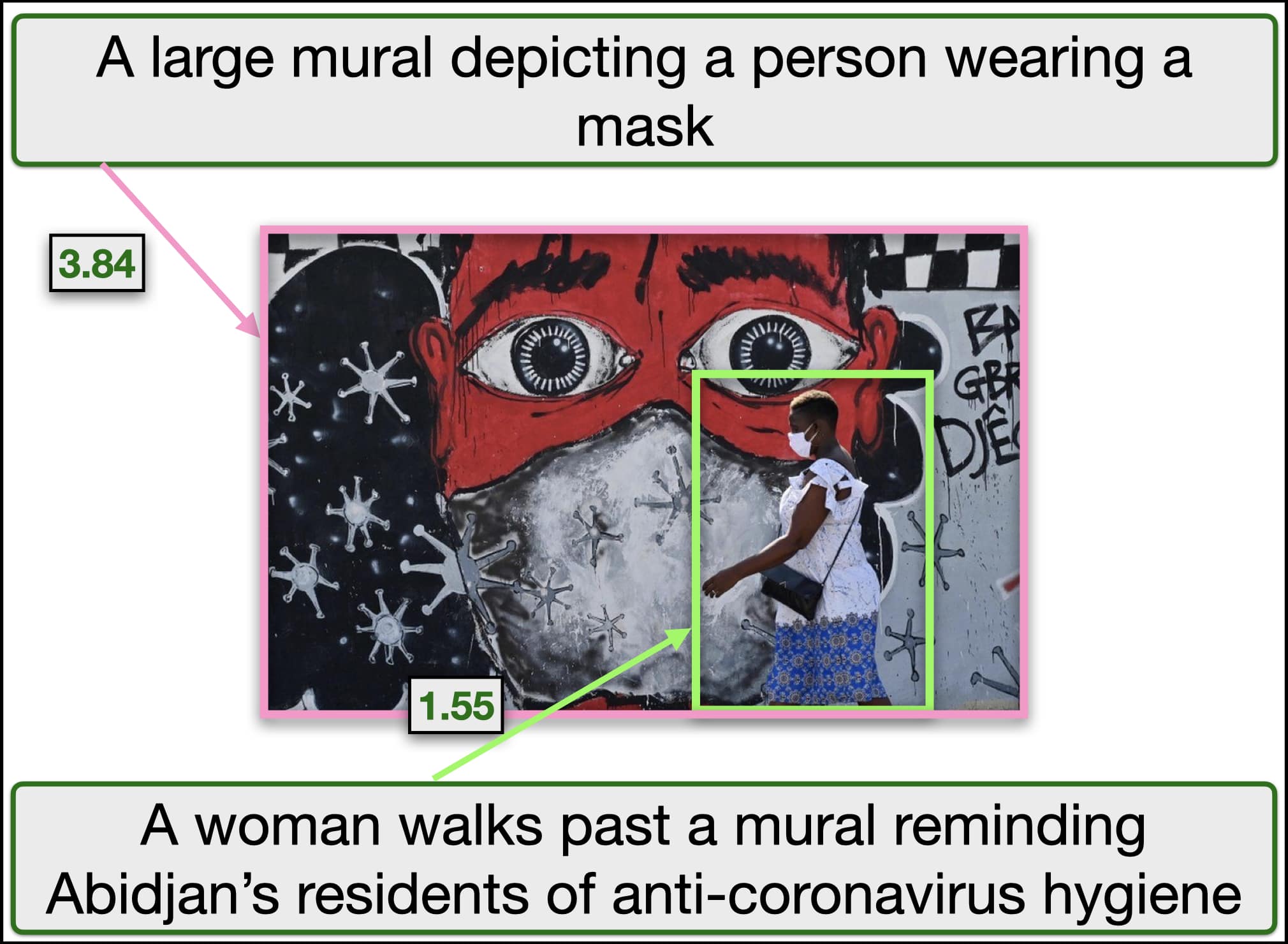} }}%
    \hfill
    \end{center}
    \begin{center}
    \subfloat{{\includegraphics[width=0.32\linewidth]{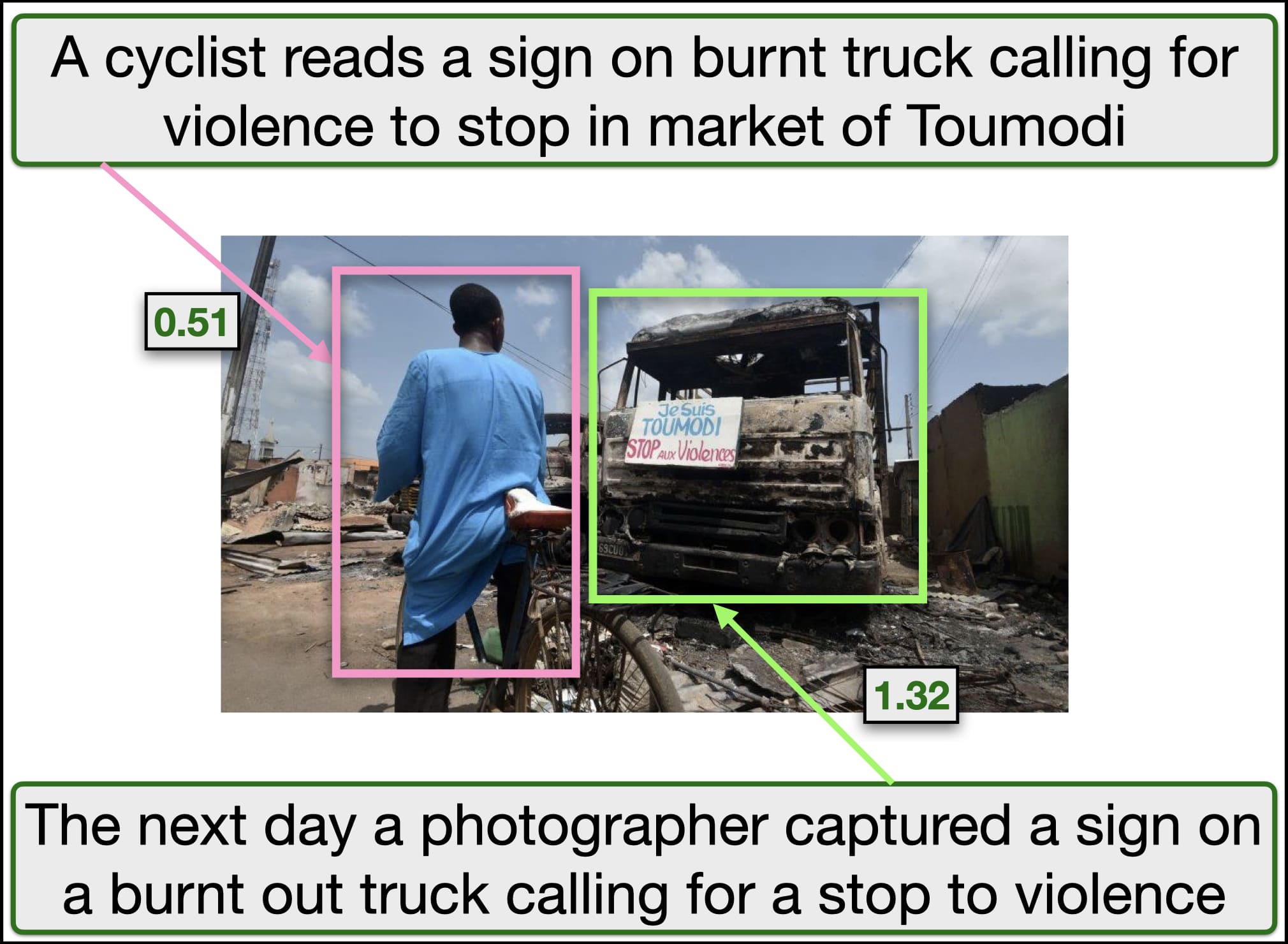} }}%
    \hfill
    \subfloat{{\includegraphics[width=0.32\linewidth]{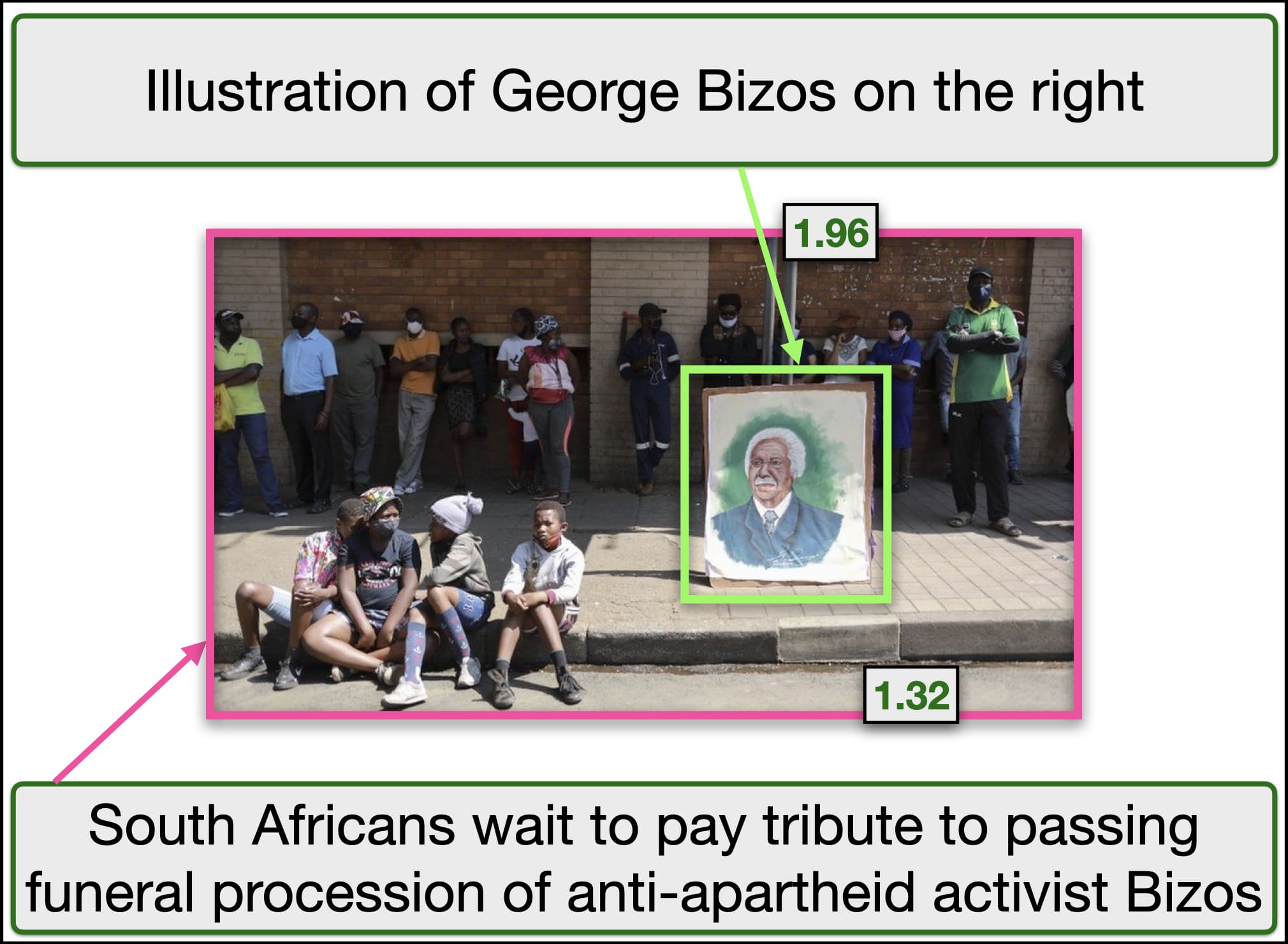} }}%
    \hfill
    \subfloat{{\includegraphics[width=0.32\linewidth]{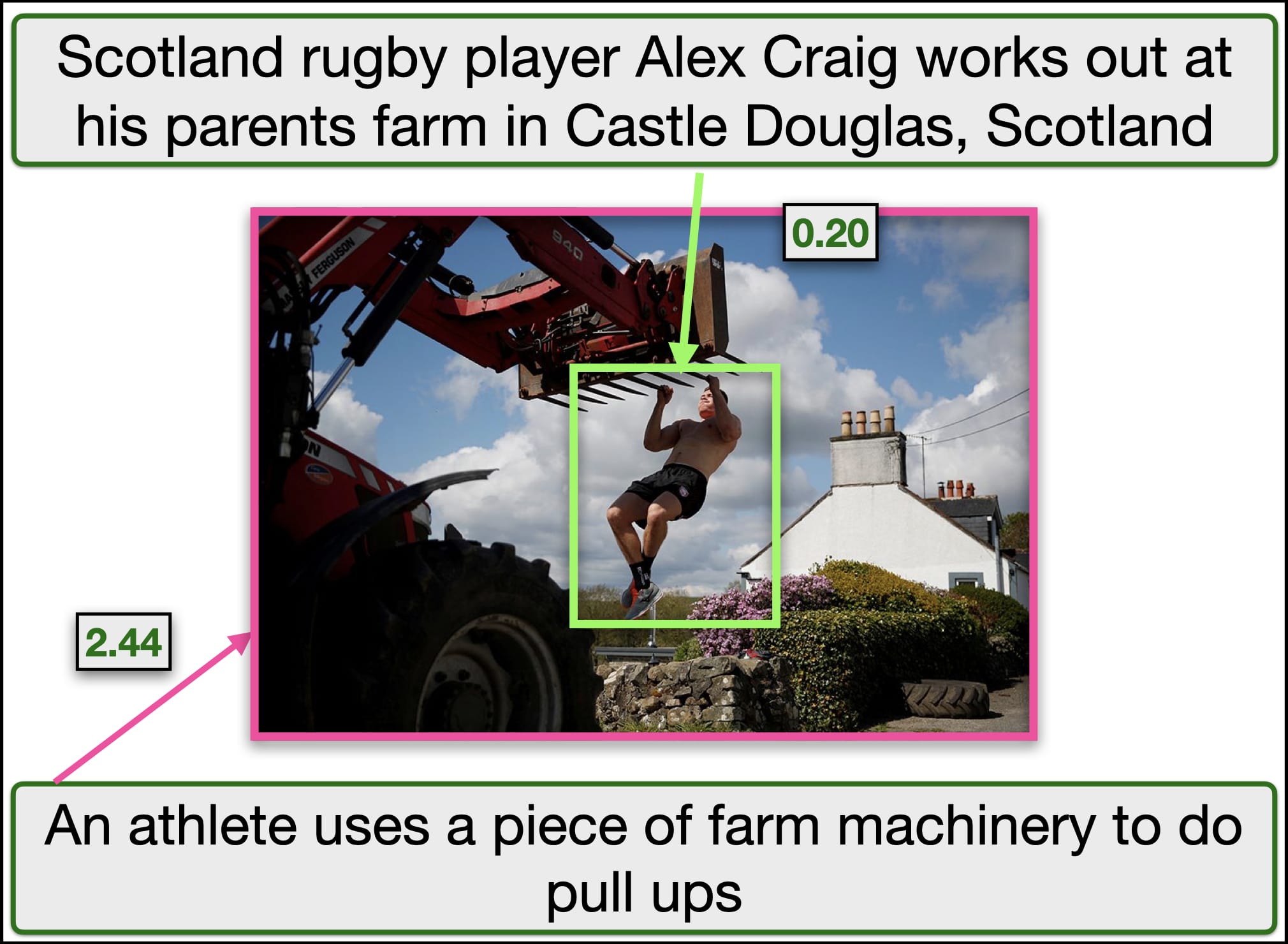} }}%
    \hfill
\end{center}
\begin{center}
    \subfloat{{\includegraphics[width=0.32\linewidth]{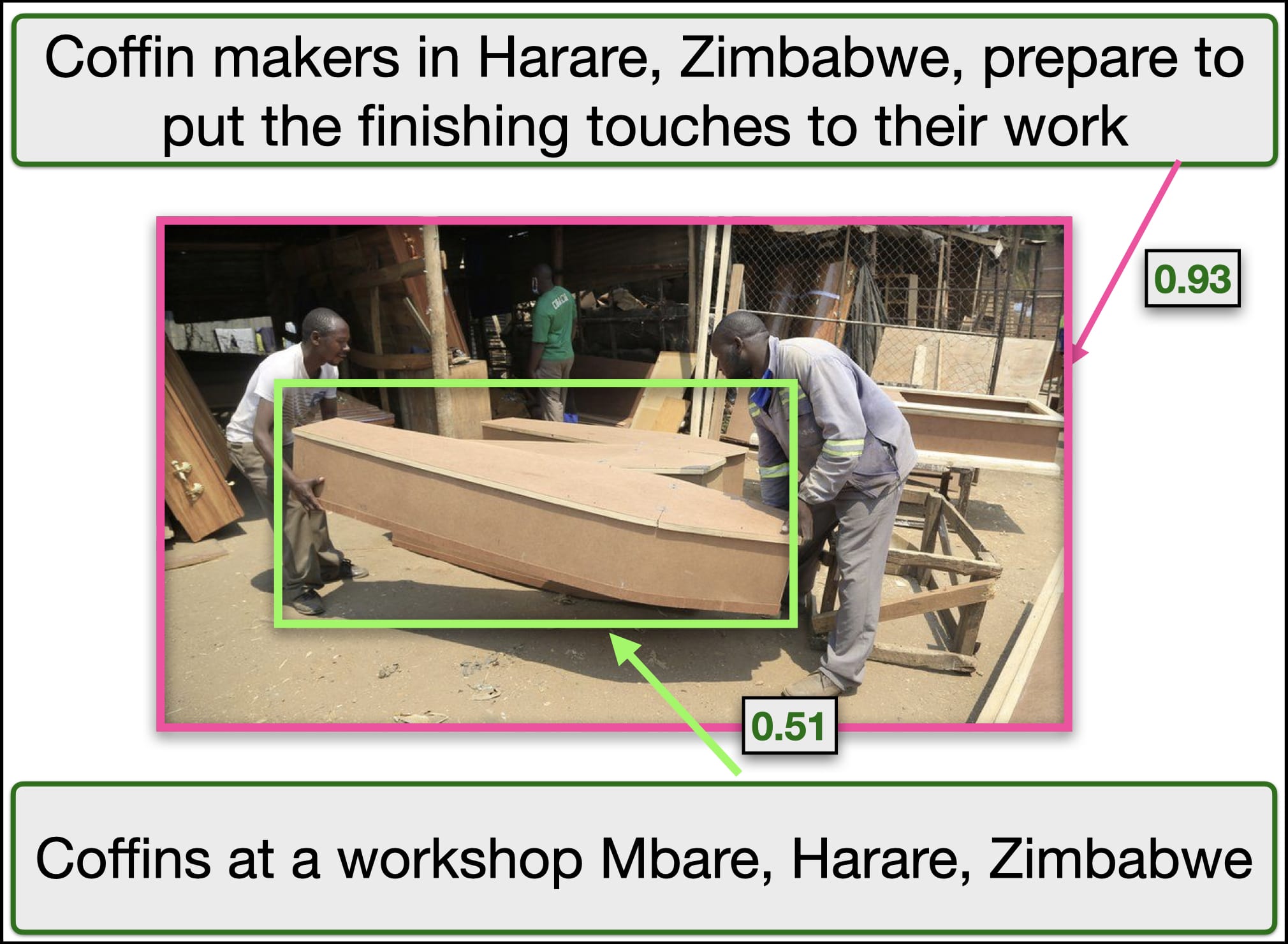} }}%
    \hfill
    \subfloat{{\includegraphics[width=0.32\linewidth]{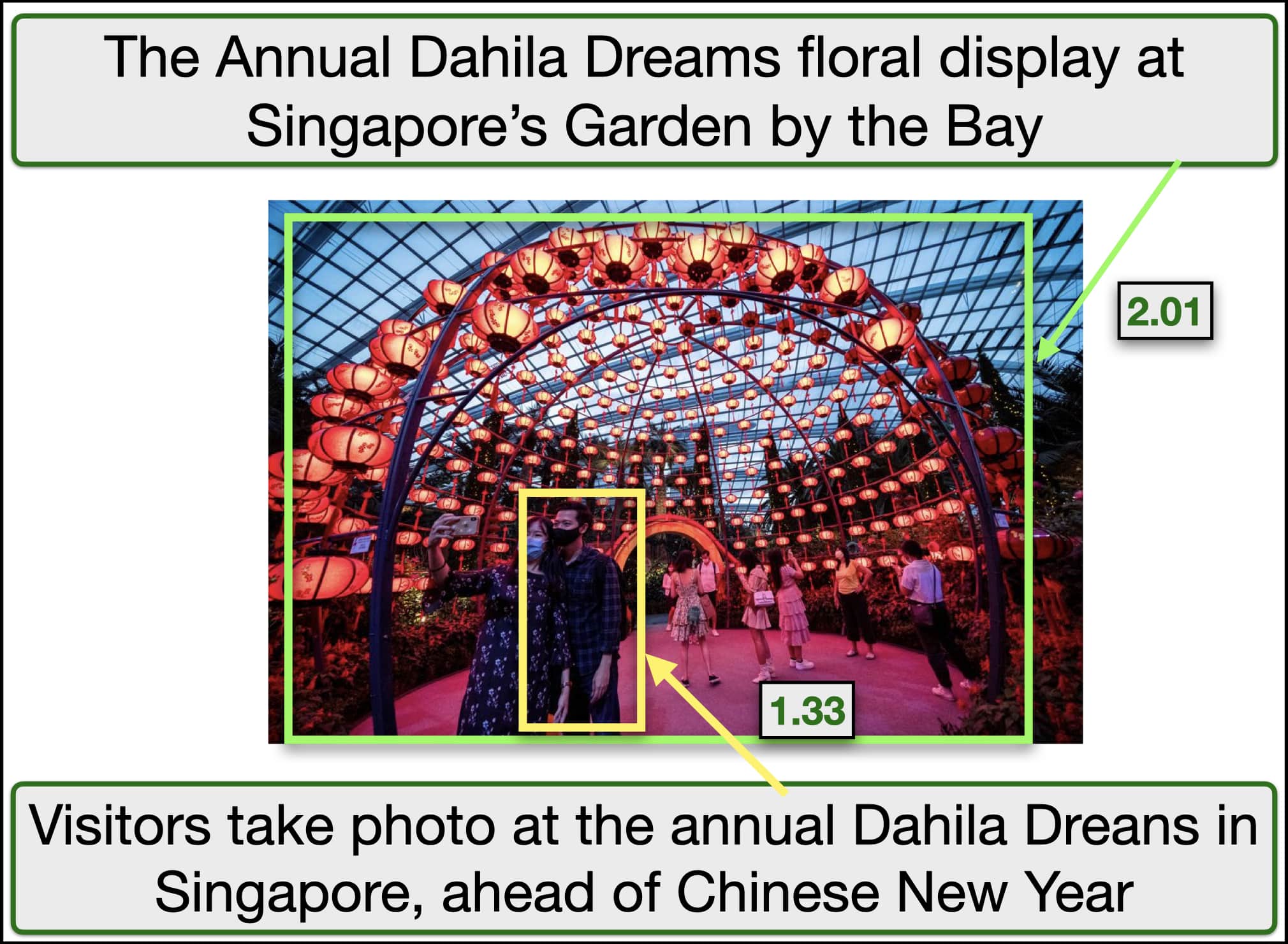} }}%
    \hfill
    \subfloat{{\includegraphics[width=0.32\linewidth]{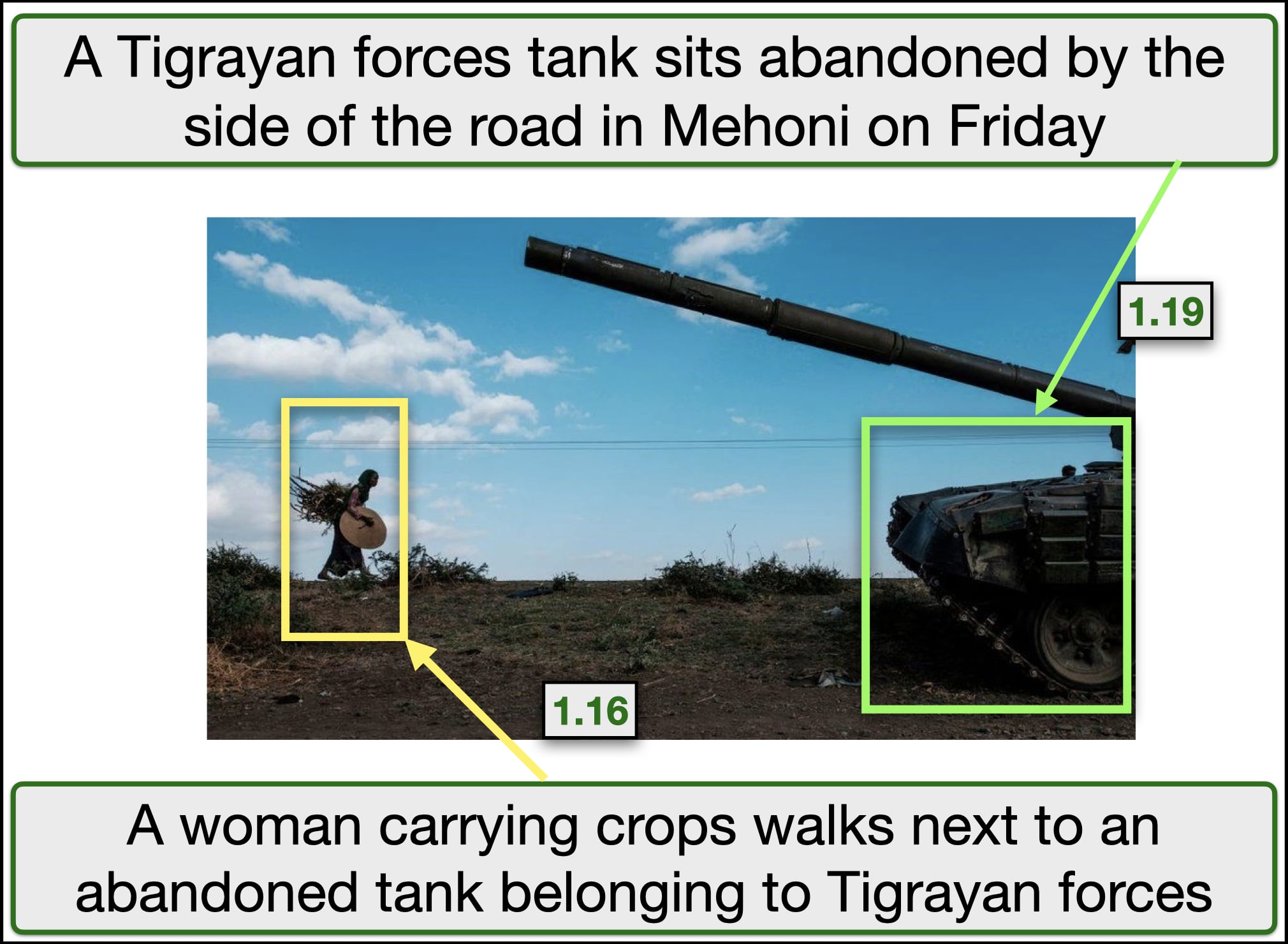} }}%
    \hfill
\end{center}
    \vspace{-0.4cm}
   \caption{Qualitative results of visual grounding of captions with the objects in the image for Not-\Ooc{} image-caption pairs. We show object-caption scores for two captions per image. The captions with \greencmd{green} border show the true captions. Scores indicate association of the most relevant object in the image with the caption.
   }
\label{fig:qualitative_res_nooc}
\end{figure*}

\section{Dataset Details}\label{sec:dataset_details}
\begin{figure*}[ht]
\begin{center}
\subfloat[Persons]{{\includegraphics[width=0.32\linewidth]{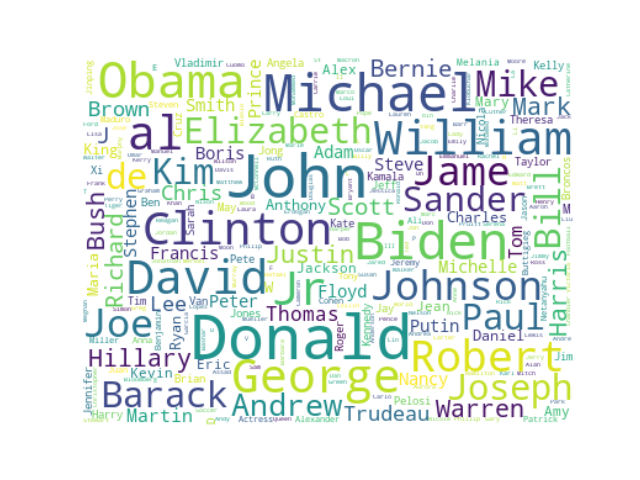}}}%
    \hfill
    \subfloat[Named Groups]{{\includegraphics[width=0.32\linewidth]{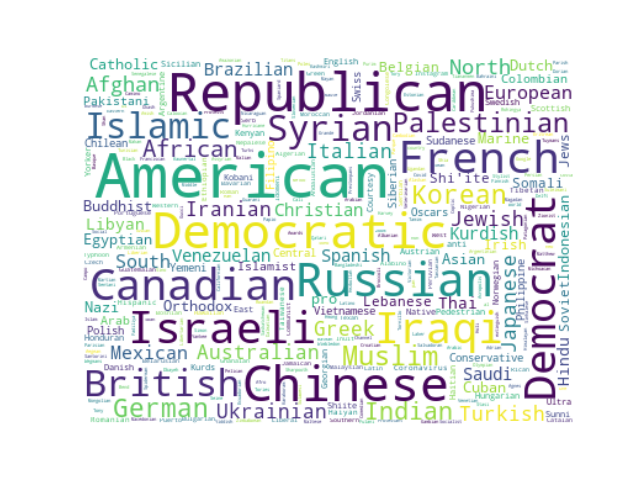} }}%
    \hfill
    \subfloat[Named Facilities]{{\includegraphics[width=0.32\linewidth]{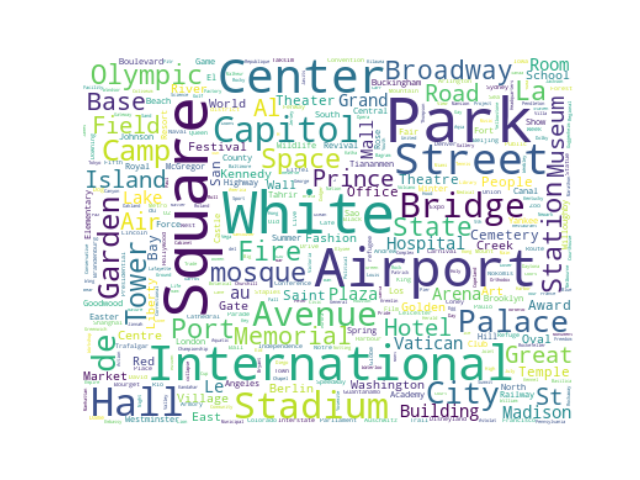} }}
    \end{center}
    
    \begin{center}
    \subfloat[Geopolitical entities]{{\includegraphics[width=0.32\linewidth]{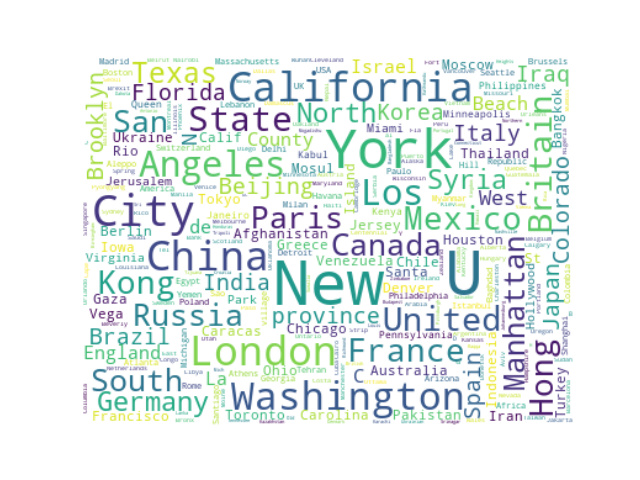} }}%
    \hfill
    \subfloat[Named Locations]{{\includegraphics[width=0.32\linewidth]{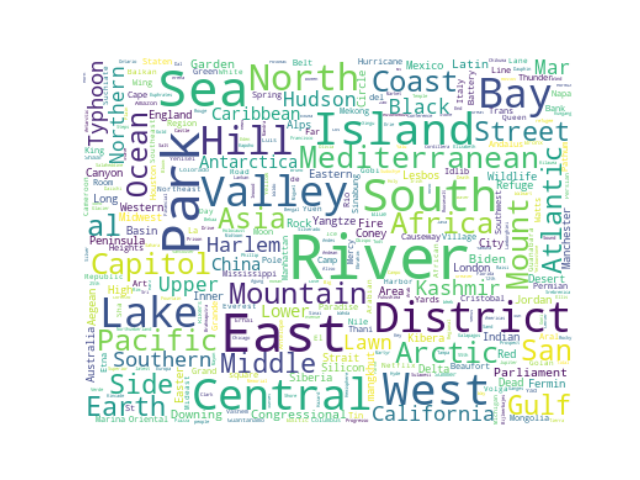} }}%
    \hfill
    \subfloat[Named events]{{\includegraphics[width=0.32\linewidth]{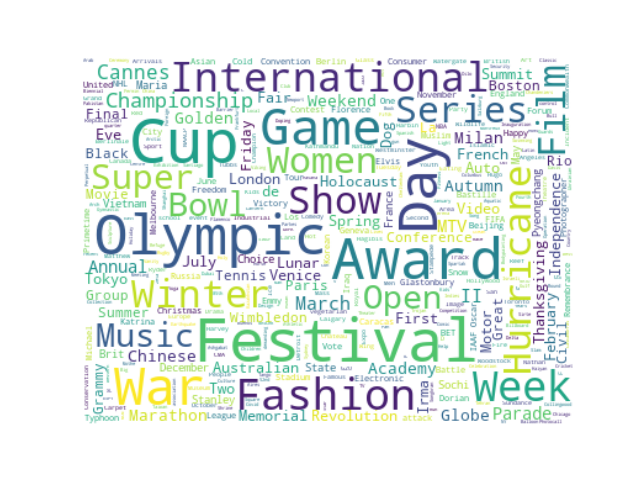} }}%
    \hfill
    \end{center}
    \begin{center}
    \subfloat[Named Organizations]{{\includegraphics[width=0.32\linewidth]{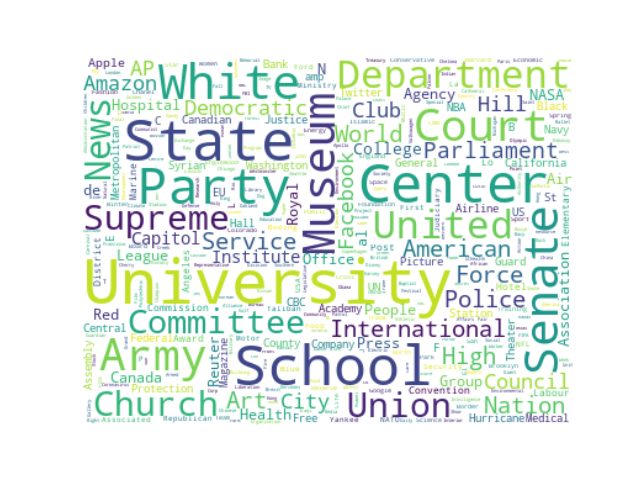} }}%
    \hfill
    \subfloat[Work Of Art]{{\includegraphics[width=0.32\linewidth]{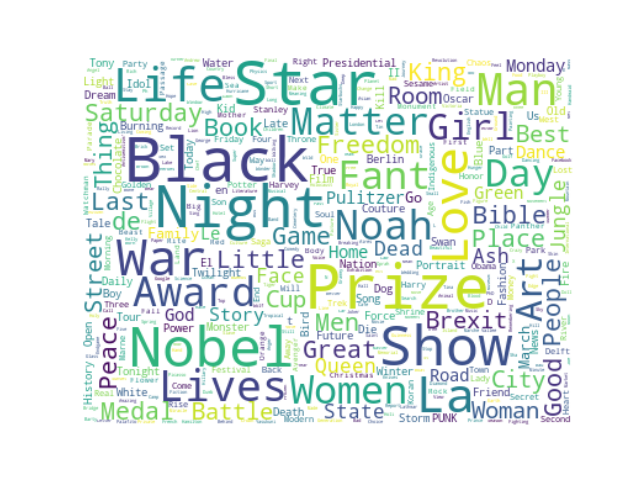} }}%
    \hfill
    \subfloat[Time]{{\includegraphics[width=0.32\linewidth]{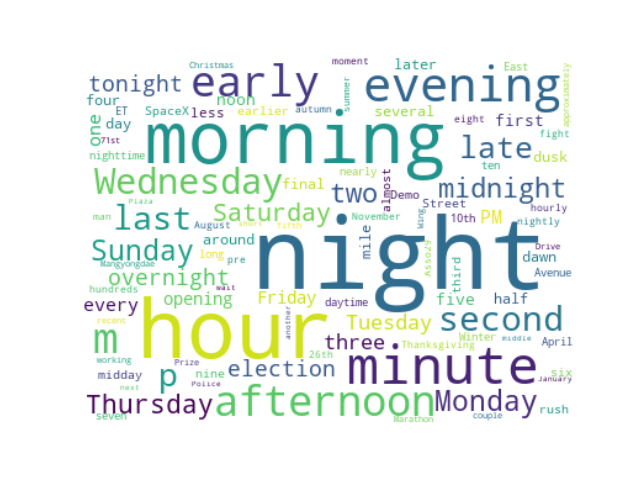} }}%
    \hfill
\end{center}
   \caption{Figure shows Word Clouds corresponding to different named entities used for text pre-processing for the experiments conducted in the main paper. Persons (a) shows names of the person; Named Groups (b) include Nationalities / religious / political groups; Named Facilities (c) include Buildings, airports, highways, bridges, etc; Geopolitical entities (d) include Countries, Cities, States;  Named Locations (e) include locations that are not geo-political entities like mountain ranges, bodies of water; Named events (f) include certain famous events like like battles, wars, sports events; Named Organizations (g) include names of Companies, agencies, institutions; Work Of Art (h) includes Titles of books, songs, awards, etc; Time (i) includes times smaller than a day   
   }
\label{fig:word_clouds}
\end{figure*}

We build a web-based annotation tool based on Python Flask and MySQL database to cleanup and label the samples of our dataset. 
We first use Google Cloud Translate API\footnote{\label{translate}https://cloud.google.com/translate} library based on Google's language detection to translate non-English captions to English. %
We then cleanup the captions by removing image/source credits by using our web tool to avoid over-fitting to specific news sources. 
Finally, we manually annotate \COUNTVALIDATION{} pairs from our Test Set. Note that the images used in our test split are gathered from Fact-Checks section of the Fact-Checking website Snopes\footnote{\label{}https://www.snopes.com/fact-check/} and from several other news websites. We ensure equal split of both \Ooc{} and Not-\Ooc{} images in the Test split for a fair evaluation.
For every image, up to 4 caption pairs with diverse vocabulary are annotated. 
Duplicate caption pairs are removed during annotation from Test split. 
However, we did not clean up up duplicates from the training set. 
Several samples from our dataset along with the corresponding captions from different sources are shown in Fig~\ref{fig:dataset_samples}. Then in Figure~\ref{fig:qualitative_res_ooc} and Figure~\ref{fig:qualitative_res_nooc} we visualize the image-caption groundings learnt by our the trained model. Word Clouds corresponding to different named entities in the captions from the dataset are shown in Figure~\ref{fig:word_clouds}. Table~\ref{tab:named_entities} lists number of captions from the dataset that contain these entities.

\begin{table}[ht]
\begin{center}
\begin{tabular}{c|c}
\hline
\toprule
Named Entity & \% of Captions   \\
\midrule
Person & 0.74 \\
Named Groups & 0.20 \\
Geopolitical entity & 0.90\\
Named Locations & 0.06 \\
Named Events & 0.05 \\
Named Organizations & 0.48 \\
Work-Of-Art & 0.05 \\
Time & 0.02 \\
\bottomrule
\end{tabular}
\vspace{-0.4cm}
\end{center}
\caption{The table documents total percentage of captions from the dataset that contain each of the named entities.}
\label{tab:named_entities}
\end{table}

\section{Setup \& Evaluation Metrics}\label{sec:setup}

\noindent
\textbf{Experimental Setup:} 
Our dataset consists of 160K training, 40K validation and \COUNTVALIDATION{} test images.  Every image is associated with one or more captions in the training set and only 2 captions in the test set. 
During training, we do not make use of \ooc\ labels. 
We train our model only with \textit{match} vs \textit{no match loss} and evaluate for \ooc\ image detection using the algorithm proposed in main paper. 
For all the experiments, we use a threshold value $t = 0.5$, to compute IoU overlap for image regions as well as for textual similarity.
\\
\\
\noindent
\textbf{Hyperparameters:} 
We run all our experiments on a Nvidia GeForce GTX 2080 Ti card. 
We use a learning rate of 1e-3 with Adam optimizer for our model with decay when validation loss plateaus for 5 consecutive epochs, early stopping with a patience of 10 successive epochs on validation loss. 
For the baselines, we use their default hyperparameters used in their original papers.

\medskip
\noindent
\textbf{Quantitative Results:} 
In the main paper, we evaluate our method using three metrics explained below:

\begin{itemize}
\item[(1)] \textit{Match Accuracy} quantifies how well the caption grounds with the objects in the image. 
To compute this numerically, we pair every image $I$ in the dataset with a matching caption $C_{m}$ and a random caption $C_{r}$ and 
compute scores for both the captions with the image. 
A higher score for matching caption ($s_m$) compared to random caption ($s_r$), i.e., $s_m > s_r$ indicates correct prediction.
\item[(2)] \textit{Object IoU} evaluates how well the caption grounds with the correct object in the image. 
For example, if the caption is ``a woman buying groceries'', the caption should ground with particular ``woman'' in the image.
For this, we select the object with maximum score 
and compare it with GT object (provided in the dataset) using IoU overlap.  
Note that we do not have these annotations for our dataset, hence we evaluate this metric on the RefCOCO~\cite{yu2016modeling} dataset. 
\item[(3)] \textit{\Ooc{} (OOC) Accuracy} evaluates the model on the out-of-context classification task described in main paper %
We collected an equal number of samples for both the classes (\Ooc\ and Not-\Ooc) for fair evaluation.
\end{itemize}

\section{Additional Experiments}\label{sec:extra_experiments} 
\subsection{Does Data Augmentation Help?}
To mitigate over-fitting and improve the overall performance of our model, we apply several augmentations during training.
For each of the detected objects  in the image, we apply color jitter by varying the hue and saturation by a factor of 0.2 and applied random horizontal flip and random rotate (by angle of 10 degrees). Note that we do not apply these augmentations during inference. And for each of the captions, we pre-process them by replacing named entities with the corresponding hypernyms as shown in the main paper. Text pre-processing is always applied to all the captions in the dataset for all the splits. We analyze the effect of each of these augmentations in Table~\ref{tab:aug_effect}. 
\begin{table}[ht]
\begin{center}
\begin{tabular}{c|c}
\hline
\toprule
Augmentation & Context Acc.  \\
\midrule
J & 0.73 \\
J + R & 0.74 \\
NER & 0.78\\
J + R + NER & \textbf{0.85} \\
\bottomrule
\end{tabular}
\vspace{-0.4cm}
\end{center}
\caption{The table illustrates the effect applying different augmentations during training. J (Jitter) and R (Random Rotate) are applied on detected objects and NER (Named Entity Recognition Replacement) is applied on textual caption. We notice that applying all these augmentations together gives us the best performing model}
\label{tab:aug_effect}
\end{table}

\medskip
\noindent
\subsection{How to Encode Text?}
We evaluate which of the two text encodings (word-level or text-level) learn better object-caption groundings. For word-level encoding, we compute fixed-size embeddings for every word in the caption and combine with every detected object during experiments. For text-level encoding, a single embedding is generated for the entire caption and combined with detected objects in the image. Results are shown in Table~\ref{tab:word_vs_text}. We notice that encoding the entire caption as single embedding consistently outperforms word-level encodings given that text-level embeddings contain richer information about certain object attributes like relative positions thereby learning better visual grounding. Hence, we used text-embeddings for all our experiments in the main paper. 

\begin{table}[ht]
\begin{center}
\begin{tabular}{c|c|c|c|}
\hline
\toprule
Method & Embed. Type  & Object IoU & Match Acc.  \\
\midrule
Bbox (\textit{GT}) & Word & 0.33 & 0.85 \\
 & Text & \textbf{0.36} & \textbf{0.89} \\
\hline
Bbox (\textit{Pred}) & Word & 0.20 & 0.80 \\
& Text & \textbf{0.27} & \textbf{0.88} \\
\bottomrule
\end{tabular}
\vspace{-0.4cm}
\end{center}
\caption{Ablations study of word vs sentence embeddings on RefCOCO dataset~\cite{yu2016modeling}. Text is embedded using Glove~\cite{pennington2014glove} pre-trained embeddings. \textit{GT} indicates Ground Truth bounding boxes used for experiments. \textit{Pred} indicates bounding boxes predicted by Mask-RCNN used for experiments.}
\label{tab:word_vs_text}
\end{table}

\subsection{Do Object-Detector Features Help?}
We experiment if same backbone network (ResNet-50~\cite{resnet_18}) trained on different tasks (Image Classification and Object Detection) can help learn better object-level features ultimately learning better object-caption groundings. Results are shown in Table~\ref{tab:tab2}. We notice that Mask-RCNN pre-trained backbone performs better in comparison to ImageNet pre-trained backbone verifying that object detector features help learn better object-caption groundings. Hence, we used Mask-RNN pre-trained backbone as image encoder for our experiments in the main paper.

\begin{table}
\begin{center}
\begin{tabular}{l|c|c|c}
\hline
Object Features & Bbox & Object IoU & Match Acc. \\
\midrule
ImageNet & GT & 0.39 & 0.90 \\ 
Mask-RCNN~\cite{he2018mask} &  & \textbf{0.45} & \textbf{0.92} \\
\hline
ImageNet & Pred & 0.32 & 0.88 \\
Mask-RCNN~\cite{he2018mask} &  & \textbf{0.38} & \textbf{0.91} \\
\hline
\end{tabular}
\vspace{-0.4cm}
\end{center}
\caption{We evaluate different backbones for our self-supervised training setting.(GT) indicates we used ground truth boxes for experiments and (Pred) indicates we used bounding boxes predicted by pre-trained Mask-RCNN. On average, we have 12.6 bboxes per image (Random Chance = 0.07)
}
\label{tab:tab2}
\end{table}

\newpage

\end{document}